\documentclass{article}

\usepackage{microtype}
\usepackage{graphicx}
\usepackage{subfigure}
\usepackage{booktabs} % for professional tables
\usepackage{threeparttable}

\usepackage[accepted]{icml2024}

\usepackage{amsmath}
\usepackage{amssymb}
\usepackage{mathtools}
\usepackage{amsthm}

\theoremstyle{plain}

\theoremstyle{definition}

\theoremstyle{remark}

\usepackage[textsize=tiny]{todonotes}

\usepackage[utf8]{inputenc} % allow utf-8 input
\usepackage[T1]{fontenc}    % use 8-bit T1 fonts
\usepackage{url}            % simple URL typesetting
\usepackage{booktabs}       % professional-quality tables
\usepackage{amsfonts}       % blackboard math symbols
\usepackage{nicefrac}       % compact symbols for 1/2, etc.
\usepackage{microtype}      % microtypography
\usepackage{xcolor}         % colors
\usepackage{wrapfig}
\usepackage[colorlinks,linkcolor=magenta,filecolor=blue,citecolor=blue,urlcolor=blue]{hyperref}%
\usepackage[capitalize,noabbrev]{cleveref}

\usepackage{bbm}
\usepackage[ruled]{algorithm2e}
\usepackage{algpseudocode}
\usepackage{authblk}
\usepackage{thmtools,thm-restate}
\usepackage{multirow}
\usepackage{subcaption}
\usepackage{pifont}
\usepackage{makecell}
\usepackage{multicol}
\usepackage{enumitem}
\newcommand{\kivi}[1]{\texttt{KIVI}}
\newcommand{\kivitwo}[1]{\texttt{KIVI-2}}
\newcommand{\kivifour}[1]{\texttt{KIVI-4}}
\usepackage{amsmath,amsfonts,bm}

\def\eqref#1{equation~(\ref{#1})}
\def\Eqref#1{Equation~(\ref{#1})}
\def\1{\bm{1}}

\def\vt{{\bm{t}}}

\def\mA{{\bm{A}}}

\def\mW{{\bm{W}}}
\def\mX{{\bm{X}}}

\DeclareMathAlphabet{\mathsfit}{\encodingdefault}{\sfdefault}{m}{sl}
\SetMathAlphabet{\mathsfit}{bold}{\encodingdefault}{\sfdefault}{bx}{n}

\icmltitlerunning{\kivi{}: A Tuning-Free Asymmetric 2bit Quantization for KV Cache}

\begin{document}

\twocolumn[
\icmltitle{\kivi{}: A Tuning-Free Asymmetric 2bit Quantization for KV Cache}



\icmlsetsymbol{equal}{*}

\begin{icmlauthorlist}
\icmlauthor{Zirui Liu}{equal,rice}
\icmlauthor{Jiayi Yuan}{equal,rice}
\icmlauthor{Hongye Jin}{tamu}
\icmlauthor{Shaochen (Henry) Zhong}{rice}\\
\icmlauthor{Zhaozhuo Xu}{stevens}
\icmlauthor{Vladimir Braverman}{rice}
\icmlauthor{Beidi Chen}{cmu}
\icmlauthor{Xia Hu}{rice}
\end{icmlauthorlist}

\icmlaffiliation{rice}{Rice University}
\icmlaffiliation{tamu}{Texas A\&M University}
\icmlaffiliation{stevens}{Stevens Institute of Technology}
\icmlaffiliation{cmu}{Carnegie Mellon University}


\icmlcorrespondingauthor{Zirui Liu}{zl105@rice.edu}
\icmlcorrespondingauthor{Jiayi Yuan}{jy101@rice.edu}

\icmlkeywords{Machine Learning, ICML}

\vskip 0.3in
]



\printAffiliationsAndNotice{\icmlEqualContribution\!\!. The order of authors is determined by flipping a coin.} % otherwise use the standard text.

\begin{abstract}

Efficiently serving large language models (LLMs) requires batching of many requests to reduce the cost per request. Yet, with larger batch sizes and longer context lengths, the key-value (KV) cache, which stores attention keys and values to avoid re-computations, significantly increases memory demands and becomes the new bottleneck in speed and memory usage. Additionally, the loading of the KV cache causes the computational core to be idle, which limits the inference speed.
A straightforward and effective solution to reduce KV cache size is quantization, which decreases the total bytes taken by KV cache. However, there is a lack of in-depth studies that explore the element distribution of KV cache to understand the hardness and limitation of KV cache quantization. 
\quad
To fill the gap, we conducted a comprehensive study on the element distribution in KV cache of popular LLMs. Our findings indicate that the key cache should be quantized per-channel, i.e., group elements along the channel dimension and quantize them together. In contrast, the value cache should be quantized per-token. From this analysis, we developed a tuning-free 2bit KV cache quantization algorithm named \kivi{}. With hardware-friendly implementation, \kivi{} can enable Llama, Falcon, and Mistral models to maintain almost the same quality while using $\mathbf{2.6\times}$ less peak memory (including model weight). This reduction in memory usage enables up to $\mathbf{4\times}$ larger batch size, bringing $\mathbf{2.35\times \sim 3.47\times}$ throughput on real LLM inference workload. The source code is available at \url{https://github.com/jy-yuan/KIVI}.

\end{abstract}

\section{Introduction}

Large Language Models (LLMs) have demonstrated strong performance across a wide range of tasks \citep{gpt3,taylor2022galactica,yuan2023large,chuang2024understanding}.
However, their deployment is very costly, requiring a large number of hardware accelerators such as GPUs.
Given these substantial costs, one natural way to reduce the cost per request is to combine a sufficient number of requests together for batch processing.
However, in this batch inference scenario, the key-value cache (KV cache), which holds the attention keys and values during generation to prevent re-computations, is becoming the new memory and speed bottleneck.
This bottleneck becomes more pronounced with larger batch sizes and longer context lengths. For instance, in 540B PaLM, with a batch size of 512 and a context length of 2048, KV cache alone can take 3TB. This is 3 times the size of the model's parameters \citep{pope2023efficiently}.
Also, the GPU SRAM has to load the whole KV cache from the GPU device memory for every token generated, during which the computational cores are idle.
Thus, reducing KV cache size in LLMs while maintaining accuracy is important.

Existing works towards this problem can be roughly divided
into three categories.
First, some work suggests reducing the number of heads in KV cache, such as multi-query attention \citep{mqa} and multi-group attention \citep{gqa}.
However, these methods require either training the model from scratch or fine-tuning the existing model.
Second, another research line reduces KV cache size by evicting unimportant tokens \citep{h2o}.
Third, some other works try to solve this problem from the system perspective, e.g., offloading KV cache \citep{flexgen} or extending virtual memory and paging techniques into the attention mechanism \citep{vllm}.

\begin{figure}
 % \vspace{-4em}
  \begin{center}
    \includegraphics[width=0.35\textwidth]{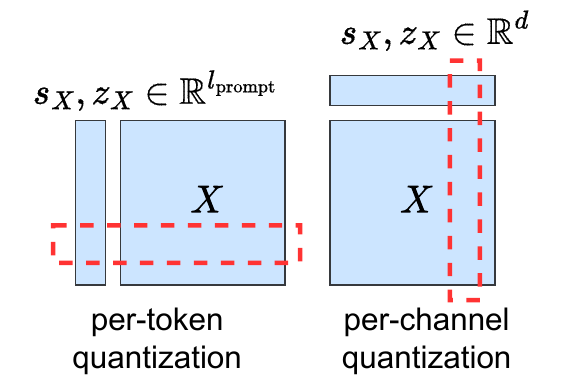}
  \end{center}
    \vspace{-1em}
  \caption{Definition of per-token and per-channel quantization.
  $\mX\in\mathbb{R}^{l_\text{prompt}\times d}$ is key/value cache where $l_\text{prompt}$ is the number of tokens and $d$ is the number of channels. $z_X$ is the zero-point, $s_X$ is the scaling factor.}
  \vspace{-1em}
  \label{fig: quant scheme}
\end{figure}

To reduce the size of KV cache, the most simple and effective way is to reduce the total bytes taken by KV cache, namely, quantization.
Unlike the well-studied weight quantization \citep{lin2023awq, xiao2023smoothquant, zhao2024atom}, to the best of our knowledge, only a few studies applied the vanilla 4bit round-to-nearest quantization to KV cache \citep{flexgen, h2o, zhao2024atom} due to the streaming nature of KV cache or other complications.  
There is a lack of in-depth studies that explore the element distribution of KV cache to understand the hardness and limitation of KV cache quantization.
To fill the gap, we study the element distribution of KV cache.
Our analysis suggests: 
\begin{itemize}
    \item For key cache, there are a few fixed channels whose magnitudes are very large, which is consistent with previous finding \citep{lin2023awq, xiao2023smoothquant}.
    Thus, as shown in Figure \ref{fig: quant scheme} right, key cache should be quantized per-channel, i.e., group elements along the channel dimension and quantize them together.
    In this way,
   it can confine the error to each individual channel, without impacting the other normal channels.
     \item For value cache, there is no obvious outlier pattern.
     Although value cache has no obvious outlier pattern, we experimentally show that it can only be quantized per-token because it is used to calculate the attention output, which is essentially a value cache mixer.
     As shown in Figure \ref{fig: quant scheme} left,
     the per-token quantization can confine the error inside each individual token and ensure that the quantization of one token does not adversely impact the others.

\end{itemize}

Based on the above insights, we propose \kivi{}, a plug-and-play extreme low-bit KV cache quantization method.
\kivi{} quantizes key cache per-channel and quantizes value cache per-token.
The per-token value cache quantization aligns well with the streaming nature of auto-regressive inference, allowing newly quantized tensors to be directly appended to the existing quantized value cache by token dimension.
However, for per-channel key cache quantization, the quantization process spans different tokens, which cannot be directly implemented in this streaming setting.
Since the number of tokens in key cache can be arbitrary, 
our key idea is to split key cache into two parts.
The \textbf{first} part is the grouped key cache, which contains several groups of tokens and each group has a certain number of tokens.
The \textbf{second} part is the residual key cache, which does not have a sufficient number of tokens to form a complete group.
Similarly, we split value cache into the grouped and residual parts to maintain the accuracy.
We only apply group-wise quantization to the grouped key cache and value cache, while the residual key cache and value cache are kept in full precision.
The grouped and residual parts can be combined using tiled matrix multiplication when computing attention scores.
Our contributions are summarized as follows:

\begin{itemize}
    \item \textbf{Extensive analysis regarding the outlier patterns and quantization error of KV cache in commonly-used LLMs}.
    Our observations suggest that key cache should be quantized per-channel and value cache should be quantized per-token.
We also explain in depth why these caches require different quantization approaches.

     \item \textbf{A new plug-and-play 2bit KV cache quantization algorithm without any fine-tuning, \kivi{}, with hardware-friendly implementation.}
     We conduct an extensive evaluation for \kivi{} with Llama, Mistral, and Falcon on popular generation tasks.
    \kivi{} can efficiently compress KV cache to 2bit and bring $\mathbf{2.6\times}$ peak memory usage reduction for Llama-2-7B, with little to no accuracy drop.
    With our efficient system implementation, this memory reduction, in return, enables up to $\mathbf{4\times}$ larger batch size and brings $\mathbf{2.35\times \sim 3.47\times}$ throughput. 

\end{itemize}

\section{Background: Attention Inference-Time Workflow}
\label{sec: background}

The LLM attention inference-time workflow involves two phases: i) the \textit{prefill} phase, where the input prompt is used to generate KV cache for each transformer layer of LLMs; and ii) the \textit{decoding} phase, where the model uses and updates KV cache to generate the next token, one at a time. 
% At each step, the model only encodes the tokens that were created in the previous step. Moreover, each time a new token is generated, its associated key and value vectors are added to the existing KV cache. As a result, the size of the KV cache is in proportional to the total number of tokens inside the current batch. 

\paragraph{Prefill Phase.} Let $\mX \in \mathbb{R}^{b\times l_{\text{prompt}} \times d}$ be the input tensor, where $b$ is the batch size, $l_{\text{prompt}}$ is the length of the input prompt, and $d$ is the model hidden size.  
For convenience, we ignore the layer index here. 
The key, value tensors can be computed by

\vspace{-2em}
\begin{equation}
    \mX_K= \mX \mW_K, \mX_V = \mX \mW_V, \nonumber
\end{equation}
\vspace{-2em}

where $\mW_K, \mW_V\in\mathbb{R}^{d\times d}$ are the key and value layer weight, respectively.  After obtaining $\mX_K$ and $\mX_V$, they are cached in the memory for the ease of decoding.

\paragraph{Decoding Phase.} Let $\vt \in \mathbb{R}^{b\times 1 \times d}$ be the current input token embedding. 
Let $\vt_K=\vt \mW_K$ and $\vt_V=\vt \mW_V$ be the key and value layer output, respectively.
We first update KV cache:
\vspace{-1em}
\begin{align}
    \mX_K &\leftarrow \text{Concat} (\mX_K, \vt_K), \nonumber \\
    \mX_V &\leftarrow \text{Concat} (\mX_V,  \vt_V ) \nonumber,     
\end{align}
then calculate the attention output as:
\begin{align}
 \vt_Q &= \vt \mW_Q,   \nonumber \\
  \mA &= \text{Softmax} (\vt_Q \mX^{\top}_K),  \nonumber \\
 \vt_O &= \mA\mX_V,
 \label{eq: av}
\end{align}
\vspace{-2em}

where $\mW_Q$ is the weight matrix of the query layer. 
For ease of illustration, we ignore the attention output layer and the other parts of the inference workflow.

\paragraph{Memory and Speed Analysis.}
The above process is repeated until a special token indicating the sentence's conclusion is reached.
Let $l_{\textbf{gen}}$ be the number of generated tokens.
From the above analysis, the shape of KV cache is $b\times (l_{\textbf{prompt}}+l_{\textbf{gen}})\times d$.
To get a sense of the scale, consider the OPT-175B model with a batch size $b$ 512, a prompt length $l_{\textbf{prompt}}$ 512, and an output length $l_{\textbf{gen}}$ 32. The KV cache requires 1.2TB, which is 3.8 times the model weights \citep{flexgen}. 
Besides the memory, the inference speed is also decided by the KV cache size.
The GPU needs to load KV cache from GPU main memory to GPU SRAM once for every token generated during which the computational core of the chip is essentially idle \citep{pope2023efficiently, vllm}.

\section{Methodology}
% TODO1: mention the simulation setting.
% TODO2: key cache error analysis
% TODO3: more details about Table 2

In scenarios with long contexts or batched inferences, the memory and speed bottlenecks are storing and loading KV cache.
The most simple and effective way to alleviate this problem is to reduce the total bytes occupied by KV cache, specifically, quantization.
Following this motivation, we first evaluate the performance of the existing quantization method in Section \ref{sec: quant prelim}.
Our observations suggest that key and value cache should be quantized along different dimensions. We analyze the rationale behind this observation in Section \ref{sec: analysis}.
Then based on the analysis, we propose \kivi{}, a new KV cache quantization method along with its streaming data structure, detailed in Section \ref{sec: algo}.

\subsection{Preliminary Study of KV Cache Quantization}
\label{sec: quant prelim}

As we analyzed in Section \ref{sec: background}, KV cache functions as a streaming data structure, where the new tensor arrives sequentially.
Thus, optimization-based methods like GPTQ \citep{gptq} are unsuitable for quantizing KV cache due to the overhead.
To the best of our knowledge, the most flexible way for quantizing KV cache is the round-to-nearest quantization.
The $B-$bit integer quantization-dequantization process can be expressed as:

\vspace{-2em}
\begin{equation}
    Q(\mX) = \bm\lfloor \frac{\mX - z_X}{s_X} \bm\rceil, ~~~
    \mX' = Q(\mX) \cdot s_X + z_X,  \nonumber
\end{equation}
\vspace{-2em}

\begin{table}
\centering
% \vspace{-1.5em}
\small
\caption{The results of simulated KV cache group-wise quantization with various configurations. The group size is set as 32. $\mathbb{C}$ stands for per-channel quantization and $\mathbb{T}$ stands for per-token quantization. Please check the whole evaluation in Table \ref{tab:merged-lm-eval}.}
\vspace{-.5em}
\label{tab:sub_simulation}
\begin{tabular}{lcc} 
\toprule
\textbf{Llama-2-13B}                         & \multicolumn{1}{l}{\textbf{CoQA}} & \textbf{TruthfulQA}  \\ 
\midrule
16bit                               & 66.37                            & 29.53              \\
4bit (K - $\mathbb{T}$, V - $\mathbb{T}$)     & 66.48                            & 29.51              \\
\midrule
2bit (K - $\mathbb{T}$, V - $\mathbb{T}$)     & 52.93                             & 24.98              \\
2bit (K - $\mathbb{C}$, V - $\mathbb{C}$) & 2.88                            & 0.74               \\
2bit (K - $\mathbb{T}$, V - $\mathbb{C}$)   & 2.80                            & 0.26               \\
2bit (K - $\mathbb{C}$, V - $\mathbb{T}$)   & \textbf{63.53}                             & \textbf{28.60}              \\
\bottomrule
\end{tabular}
% \vspace{-3em}
\end{table}

where $z_X=\min \mX$ is the zero-point, $s_X=(\max \mX - \min \mX)/(2^B - 1)$ is the scaling factor, and $\bm\lfloor\cdot\bm\rceil$ is the rounding operation. 
Here we ignore the batch size for ease of understanding.
As shown in Figure \ref{fig: quant scheme}, $\mX$ is quantized along either the token or channel dimension group-wisely. 

\begin{figure*}[h]
    \centering
    \subfigure{
    \centering
    	\begin{minipage}[t]{0.25\linewidth}
    		\includegraphics[width=0.99\linewidth]{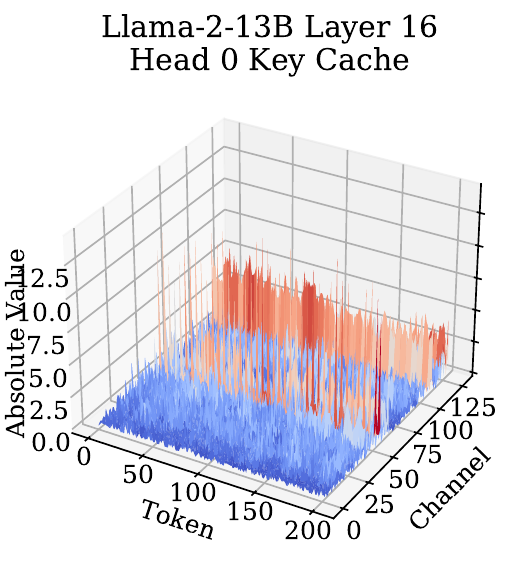}
    	\end{minipage}%
    }
    \!\!\!\!\!\!
    \subfigure{
    \centering
    	\begin{minipage}[t]{0.25\linewidth}
    		\includegraphics[width=0.99\linewidth]{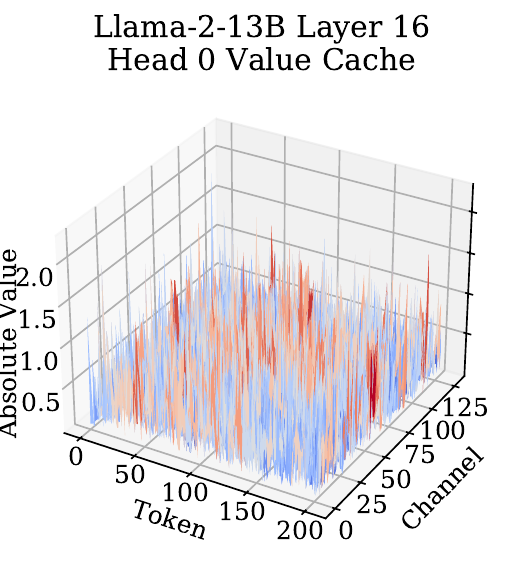}
    	\end{minipage}%
    }
    \!\!\!\!\!\!
    \subfigure{
    \centering
    	\begin{minipage}[t]{0.25\linewidth}
    		\includegraphics[width=0.99\linewidth]{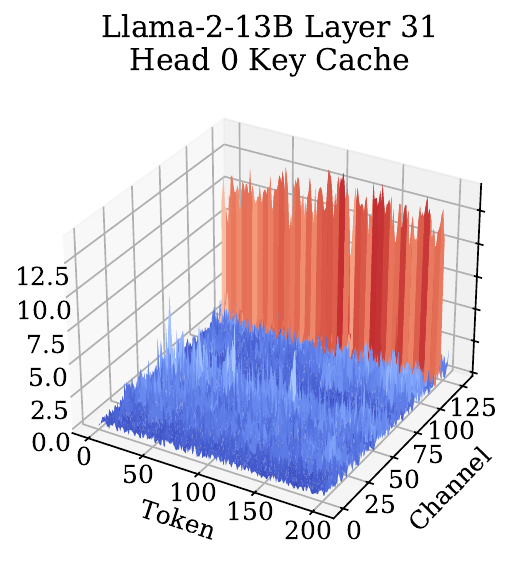}
    	\end{minipage}
    }
    \!\!\!\!\!\!
    \subfigure{
    \centering
    	\begin{minipage}[t]{0.25\linewidth}
    		\includegraphics[width=0.99\linewidth]{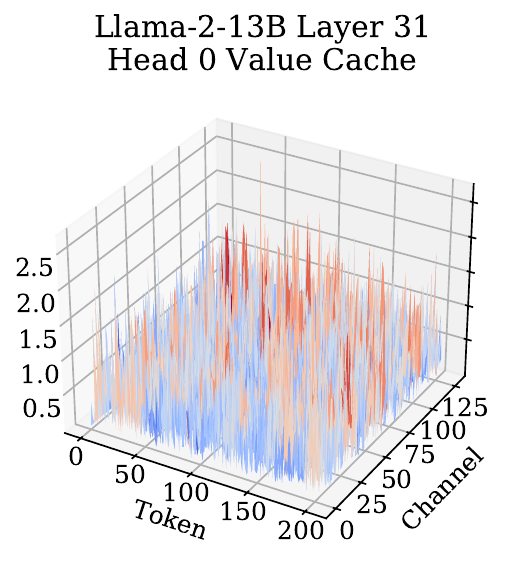}
    	\end{minipage}
    }
    \!\!\!\!\!\!
    \subfigure{
    \centering
    	\begin{minipage}[t]{0.25\linewidth}
    		\includegraphics[width=0.99\linewidth]{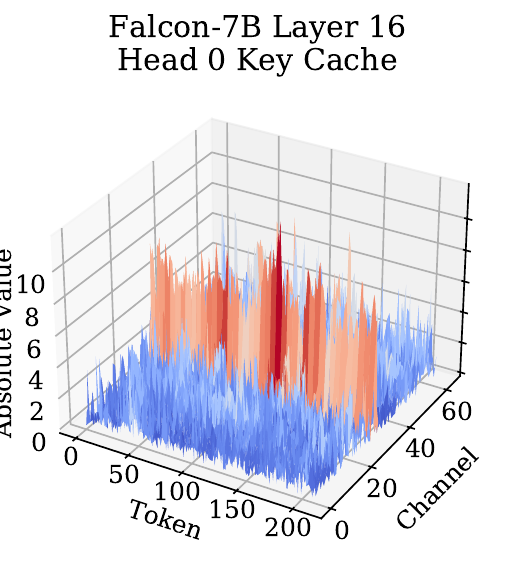}
    	\end{minipage}%
    }
    \!\!\!\!\!\!
    \subfigure{
    \centering
    	\begin{minipage}[t]{0.25\linewidth}
    		\includegraphics[width=0.99\linewidth]{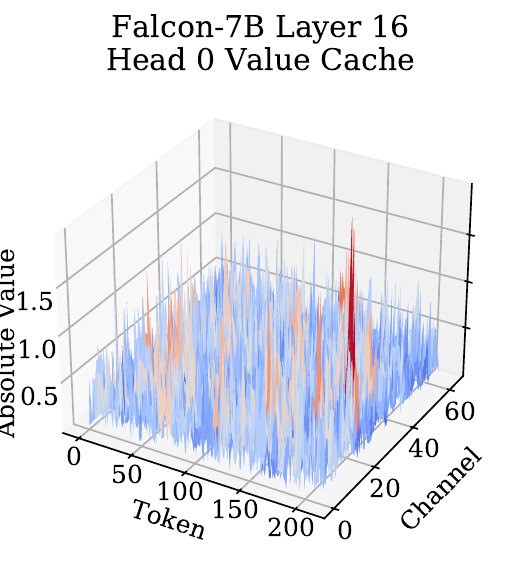}
    	\end{minipage}%
    }
    \!\!\!\!\!\!
    \subfigure{
    \centering
    	\begin{minipage}[t]{0.25\linewidth}
    		\includegraphics[width=0.99\linewidth]{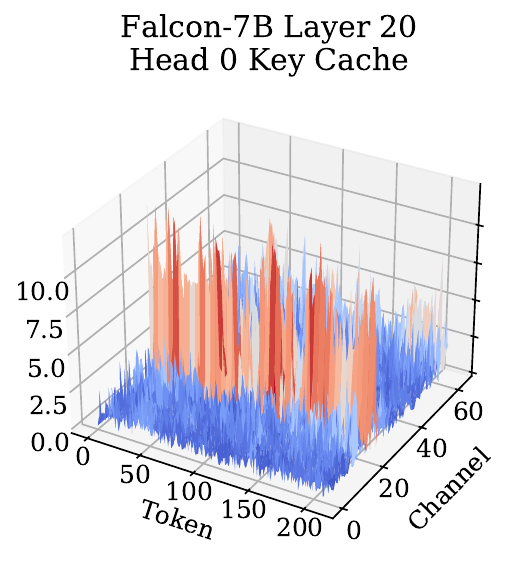}
    	\end{minipage}
    }
    \!\!\!\!\!\!
    \subfigure{
    \centering
    	\begin{minipage}[t]{0.25\linewidth}
    		\includegraphics[width=0.99\linewidth]{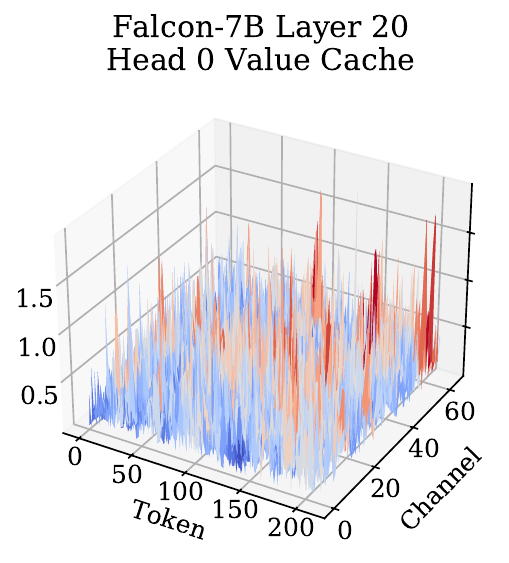}
    	\end{minipage}
    }
    \vspace{-1em}
    \caption{Magnitude of key and value cache for Llama-2-13B and Falcon-7B. We observe (1) for key cache, there are a few channels whose magnitudes are very large.
    (2) for value cache, there is no obvious outlier pattern.}
    \label{fig: vis}
\end{figure*}

Considering the streaming nature of KV cache, previous studies often apply per-token quantization to both key and value cache since the newly quantized KV cache can be naively added to the existing quantized one along the token dimension \citep{flexgen}. While per-channel quantization is non-trivial, we have designed a padding method to implement per-channel quantization to explore its effect on both key and value cache.

\paragraph{Setting.}
In Table \ref{tab:sub_simulation}, we show the results of fake KV cache group-wise quantization with different configurations on the Llama-2-13B model for the CoQA and TruthfulQA tasks. We use a group size of 32 for all configurations.
Here fake quantization means we simulate the quantization process by first quantizing KV cache into lower precision and then dequantizing it in the attention layer.
For per-channel quantization, if the number of tokens is not divided evenly into groups, we add zero-padding to ensure it can be grouped perfectly.
In this way, we ensure that all tokens in KV cache are quantized for a fair comparison.
The detailed experimental setting can be found in Section \ref{sec: setting}.
Specifically, we observe that: 

\noindent
\textbf{OB 1.} When using the commonly used per-token quantization to both key and value caches, INT4 precision can maintain accuracy. However, reducing it to INT2 results in a notable accuracy drop.

\noindent
\textbf{OB 2.} When value cache is quantized per-channel, the accuracy significantly worsens regardless of how key cache is quantized.

\noindent
\textbf{OB 3.}  When using a lower numerical precision such as INT2, the most accurate approach is to quantize key cache per-channel and value cache per-token.

\subsection{Why Key and Value Cache Should Quantize Along Different Dimensions?}
\label{sec: analysis}
In Table \ref{tab:sub_simulation}, we observe that quantizing key cache per-channel and value cache per-token to 2bit results in a very small accuracy drop.
Here we analyze why this configuration delivers better accuracy.
In Figure \ref{fig: vis} we visualize the original KV cache distribution at different layers.
We observe that \textbf{in key cache, some fixed channels exhibit very large magnitudes, whereas in value cache, there is no significant pattern for outliers.}

\vspace{-1em}
\noindent
\paragraph{Analysis of Key Cache.}
The above observation for key cache aligns with previous findings that certain fixed columns in activations exhibit larger outliers \citep{llmint8, lin2023awq}.
The persistence of outliers within each channel means that per-channel quantization can confine the quantization error to each individual channel without impacting the other normal channels.
Thus, Figure \ref{fig: vis} explains why key cache should be quantized per-channel. 
In Table \ref{tab: v cache analysis} we show key cache relative reconstruction error $\|\frac{\mX_K-\mX_K'}{\mX_K}\|_F$, along with the relative attention score error $\|\frac{\mA-\mA'}{\mA}\|_F$ where $\mA'=\text{Softmax}(\vt_Q\mX^{'\top}_K)$. 
We observe that the per-token quantization can lead to almost $5\times$ larger attention score error than per-channel quantization, which is consistent with Figure \ref{fig: vis}.

\noindent

\begin{table}[h]
\centering
\small
% \vspace{-1em}
\caption{The relative error statistics averaged over all layers and all heads}
% \vspace{-1em}
\label{tab: v cache analysis}
\begin{tabular}{lcc} 
\toprule
  Llama-2-13B                 & K Per-Token & K Per-Channel  \\
  \midrule
Avg. $\|\frac{\mX_K-\mX_K'}{\mX_K}\|_F$          & 13.67       & 4.55           \\[2pt]
Avg. $\|\frac{\mA-\mA'}{\mA}\|_F$          & 47.00       & 9.60           \\[2pt]
Attention sparsity & \multicolumn{2}{c}{84.3\%}   \\
\bottomrule
\toprule
      & V Per-Token & V Per-Channel  \\ 
\midrule
Avg. $\|\frac{\mX_V-\mX_V'}{\mX_V}\|_F$           & 4.57       & 3.73           \\[2pt]
Avg. $\Delta$   & 3.55        & 49.89          \\ 
\bottomrule
\end{tabular}
\end{table} 

\begin{figure*}[t!]
    \centering
    \includegraphics[width=0.9\textwidth]{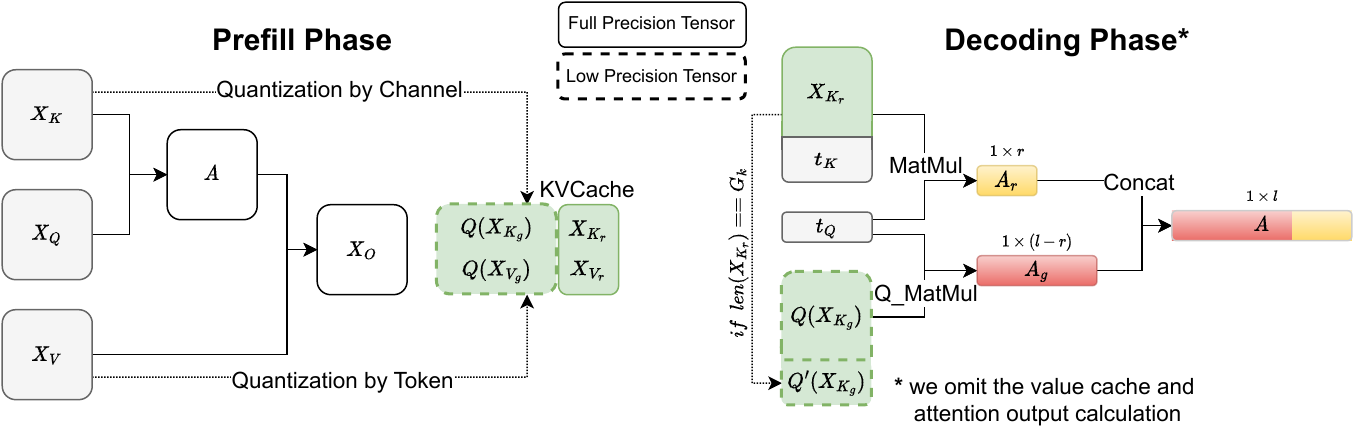}
    \vspace{-1em}
    \caption{The overview of \kivi{} algorithm. For ease of illustration, we omit the value cache and attention output parts. The detailed pseudo-code is provided in Algorithm \ref{algo: kivi}.
    Here ``Q\_Matmul'' is the mix-precision matrix multiplication which fuses the dequantization with matrix multiplication at the tiling level.}
    \label{fig: kivi}
    \vspace{-1em}
\end{figure*}

\paragraph{Analysis of Value Cache.}
Unlike key cache, value cache does not show the channel-wise outlier pattern. Furthermore, Figure \ref{fig: vis} alone cannot explain \textbf{OB2}, which indicates value cache should only be quantized per-token. 
This is because Figure \ref{fig: vis} implies that errors should be comparable for both per-token and per-channel quantization, given the absence of a clear pattern.
As shown in \Eqref{eq: av}, value cache is used to calculate the attention output $\vt_O$.
Instead of analyzing the quantization error of value cache $\mX_V$, in Table \ref{tab: v cache analysis} we analyze the relative error $\Delta=\|\frac{\mA\mX_V-\mA \mX'_V}{\mA\mX_V}\|_F$ with different quantization configurations.
Surprisingly, we observe that the per-token quantization error is almost $15\times$ smaller than per-channel quantization, which explains why \textbf{OB2} happens.
The intuition behind this observation stems from the attention sparsity.
\Eqref{eq: av} can be written as:

\vspace{-2em}
\begin{align}
    [\mA \mX_V]_{i*} = \sum_{j=1}^{l_{\text{prompt}}} \mA_{ij} [\mX_{V}]_{j*},
    \label{eq: att}
\end{align}
\vspace{-2em}

where $[\mX_{V}]_{j*}$ is the $j$-th row of $\mX_{V}$.
From \Eqref{eq: att}, the attention output is the weighted summation of value cache across various tokens, with the weights being the attention scores.
Since the attention score is highly sparse \citep{tian2023scan}, the output is just the combination of value caches of a few important tokens.
The per-token quantization can confine the error to each individual token.
Thus, quantizing other tokens does not affect the accuracy of important tokens. Consequently, per-token quantization leads to a much smaller relative error $\Delta$. 

\subsection{\kivi{}: Algorithm and System Support}
\label{sec: algo}

\paragraph{Algorithm.}
As we previously analyzed, key cache should be quantized per-channel and value cache should be quantized per-token.
Recall that key and value cache of newly generated tokens arrive sequentially.
From the implementation perspective, per-token quantization aligns well with streaming settings, allowing newly quantized tensors to be directly appended to the existing quantized value cache by token dimension.
However, for per-channel quantization, the quantization process spans across different tokens, which cannot be directly implemented in the streaming setting.
As shown in Figure \ref{fig: kivi}, our key idea to solve this problem is to group key cache every $G$ tokens and quantize them separately. 
Because the number of tokens in $\mX_K$ can be arbitrary, we split $\mX_K$ into two parts, namely, the grouped part $\mX_{K_g}=\mX_{K}[:l-r]$ and residual part $\mX_{K_r}=\mX_{K}[l-r:]$, where $l$ is the number of tokens inside the current key cache $\mX_K$, $r$ is the number of residual tokens, where $l-r$ can be divisible by $G$.

Since $\mX_{K_g}$ can be evenly divided into $(l-r)/G$ groups, we only store $Q(\mX_{K_g})$ with group-wise quantization, while $\mX_{K_r}$ is kept in full precision.
During the decoding process, each newly arrived key cache $\vt_K$ is added to $\mX_{K_r}$ and once $\mX_{K_r}$ reaches $R$ tokens, which is a hyperparameter - residual length, we quantize and concatenate it with the previously quantized $Q(\mX_{K_G})$. 
Then we reset $\mX_{K_r}$ to an empty tensor. We note that $R$ should be divisible by $G$.
With tiled matrix multiplication, the raw attention logits is then calculated as:

\vspace{-2em}
\begin{align}
\mA_g &= \vt_Q Q(\mX^{\top}_{K_g}), \nonumber \\
\mX_{K_r} &= \text{Concat}([\mX_{K_r}, \vt_K]), \nonumber \\
\mA_r &= \vt_Q \mX^{\top}_{K_r}, \nonumber \\
\mA &= \text{Concat}([\mA_g, \mA_r]).
\label{eq: attquant}
\end{align}
\vspace{-2em}

% \begin{align}
% \vt_Q \mX^{\top}_K= \text{Concat}([\vt_Q Q(\mX^{\top}_{K_g}), \vt_Q \mX^{\top}_{K_r}]).
% \label{eq: attquant}
% \end{align}

For value cache, similar to key cache, we also split it into two parts and keep the most recent value cache in full precision, namely, $\mX_{V_g}$ and $\mX_{V_r}$. Specifically, we maintain a queue and each newly arrived value cache is pushed into the queue. 
Once the queue reaches the predefined residual length $R$, the most outdated value cache is poped. Then the poped value cache is quantized per-token and concatenated with the previously quantized value cache along the token dimension. 

As shown in Figure \ref{fig: kivi}, we also emphasize that during the prefill phase, the exact key and value tensors are passed to the next layers, although only the quantized KV cache is retained in memory.
The whole algorithm can be found in Appendix \ref{app: implement} Algorithm \ref{algo: kivi}.

\paragraph{Analysis.}
In \kivi{}, the grouped key cache $\mX_{K_g}$ and value cache $\mX_{V_g}$ is quantized, while the residual key cache $\mX_{K_r}$ and value cache $\mX_{V_r}$ is kept in full precision.
By design, there are at most $R$ tokens inside $\mX_{K_r}$ or $\mX_{V_r}$. In practice, we set $R \leq 128$ and the sequence length $l_{\text{prompt}}+l_{\text{gen}}$ is often much longer than $R$.
Thus the memory overhead from $\mX_{K_r}$ and $\mX_{V_r}$ is negligible when considering the benefit from extreme low-bit quantization, especially for the long context scenarios.
Also, since the newly arrived key and value tensors are added to $\mX_{K_r}$ and $\mX_{V_r}$ in full precision, \kivi{} maintains a full precision KV cache sliding window for the local relevant tokens.
This window size is expected to be $\frac{R}{2}$ for key cache, and $R$ for value cache.
\textbf{Later in the experiment section, we show that this full precision sliding window is crucial for obtaining desirable performance on hard tasks, such as GSM8K.}

\paragraph{System Support.}
We provide a hardware-friendly implementation for running \kivi{} on GPUs.
To minimize the overhead, we have fused the dequantization process with matrix multiplication, e.g., Q\_MatMul in Figure \ref{fig: kivi}, using CUDA. 
We also implement the group-wise quantization kernel in Triton.
Our method is fully compatible with weight-only quantization.

\section{Experiments}

\subsection{Settings}
\label{sec: setting}
\paragraph{Models.} We evaluate \kivi{} using three popular model families: Llama/Llama-2~\citep{touvron2023llama, touvron2023llama2}, Falcon~\citep{penedo2023refinedweb} and Mistral~\citep{jiang2023mistral}. 
Llama and Mistral model is based on multi-head attention, while Falcon is based on multi-query attention \citep{mqa}.
We use the Hugging Face Transformers codebase and implement the \kivi{} algorithm upon it.
Following previous work \citep{flexgen}, the group size $G$ in Algorithm \ref{algo: kivi} for quantization is set as 32 across all experiments, the residual length $R$ for key and value cache is set to 128.

\paragraph{Tasks.}
As we analyzed in Section \ref{sec: background}, the KV cache size grows larger with a longer context. 
Thus, we evaluate \kivi{} under the normal context length and long context setting, respectively.
Specifically, we adopt generation tasks from LM-Eval~\citep{eval-harness} for normal context length evaluation and LongBench~\citep{bai2023longbench} for long context evaluation, respectively\footnote{The closed-end tasks such as MMLU are not ideal to evaluate \kivi{} since they only involve one decoding step and directly fetch the output logits, which is not suitable for studying the impact of compressed KV cache.}. 
For LM-eval, we adopt CoQA (Exact match accuracy), TruthfulQA (BLEU score), and GSM8K (Exact match accuracy). 
For LongBench, we chose tasks from four subgroups. 
Specifically, Qasper (F1 score) is a Single-Document QA task; QMSum (ROUGE score) and MultiNews (ROUGE score) are Summarization tasks; TREC (classification score), TriviaQA (F1 score), and SAMSum (ROUGE score) are Few-shot Learning tasks; and LCC (similarity score) and RepoBench-P (similarity score) is Code Completion task. The maximum sequence length in LongBench was set to 8192 for the Mistral model and 4096 for other models.
We also consider the needle-in-a-haystack task (NIAH) to evaluate the model's long context retrieval ability after quantizing KV cache.
Detailed NIAH setting can be found in Appendix \ref{app: niah}.

\subsection{Accuracy and Efficiency Analysis}

\subsubsection{Comparison Between Different Quantization Configurations}

We first utilize the fake quantization to demonstrate the effectiveness of our asymmetric quantization, namely, quantizing key cache per-channel and value cache per-token. 
Here fake quantization is exactly the same as in Table \ref{tab:sub_simulation}.
The results are shown in Table \ref{tab:merged-lm-eval}.
We observe that ``2bit (K per-channel, V per-token)'' consistently achieves the best results compared to all other configurations.
This is consistent with our previous analysis.
We also note that for hard generation tasks such as GSM8K, the fake ``2bit (K per-channel, V per-token)'' quantization results are significantly worse than the full precision counterparts.
However, for \kivi{} in Table \ref{tab:merged-lm-eval}, we observe that the accuracy drop is only around $2\%$ for GSM8K across different models.
As we analyzed in Section \ref{sec: algo},
the difference between fake ``2bit (K per-channel, V per-token)'' quantization and \kivi{} is that \kivi{} maintains a full precision key and value cache sliding window for the local relevant tokens.
This sliding window is crucial to maintaining accuracy for hard generation tasks such as mathematical reasoning. 

\subsubsection{Accuracy Comparison on Generation Tasks}

\begin{table}[h!]
\centering
\caption{Performance comparison between 16bit, 4-bit per-token quantization, four fake 2bit KV cache quantization similar to those in Table \ref{tab:sub_simulation}, \kivitwo{} (2bit) / \kivifour{} (4bit) across various models.
We emphasize that unlike \kivi{}, which preserves a small portion of full precision key cache $\mX_{K_r}$ and value cache $\mX_{V_r}$, all tokens in fake KV cache quantization are quantized for a fair comparison.
$\mathbb{C}$ stands for per-channel quantization and $\mathbb{T}$ stands for per-token quantization.}
\label{tab:merged-lm-eval}
\resizebox{\linewidth}{!}{
\begin{tabular}{llccc} 
\toprule
\multicolumn{2}{l}{\textbf{Model}}        & \textbf{CoQA} & \textbf{TruthfulQA} & \textbf{GSM8K}  \\
\midrule
\multirow{8}{*}{Llama-2-7B}  & 16bit  &     63.88      &         30.76        &   13.50     \\
 & 4bit (K - $\mathbb{T}$, V - $\mathbb{T}$) & 64.82 & 29.85 & 12.28 \\
\cmidrule{2-5}
 & 2bit (K - $\mathbb{C}$, V - $\mathbb{T}$)  &  59.08 & 33.10 & 5.76 \\
 & 2bit (K - $\mathbb{T}$, V - $\mathbb{T}$)     &  39.88 & 18.29 & 0.83 \\
 & 2bit (K - $\mathbb{C}$, V - $\mathbb{C}$) & 3.60 & 0.27 & 0.00 \\
 & 2bit (K - $\mathbb{T}$, V - $\mathbb{C}$)   & 1.30 & 0.49 & 0.08\\
 % & kivi-2 (w/o buffer) & 61.80 & 32.80 & 11.22 \\
 & \kivifour{} & 63.78 & 30.80 & 13.80 \\
 & \kivitwo{} &  63.05 & 33.95 & 12.74 \\
\midrule
\multirow{8}{*}{Llama-2-13B} & 16bit  &      66.37     &        29.53        &   22.67      \\
 & 4bit (K - $\mathbb{T}$, V - $\mathbb{T}$)  & 66.73 & 29.14 & 20.92 \\
\cmidrule{2-5}
 & 2bit (K - $\mathbb{C}$, V - $\mathbb{T}$)  & 63.53 & 28.60 & 12.21 \\
 & 2bit (K - $\mathbb{T}$, V - $\mathbb{T}$)   & 52.93 & 24.98 & 4.55 \\
 & 2bit (K - $\mathbb{C}$, V - $\mathbb{C}$) & 2.88 & 0.74 & 0.00 \\
 & 2bit (K - $\mathbb{T}$, V - $\mathbb{C}$)   & 2.80 & 0.26 & 0.08 \\
 % & kivi-2 (w/o buffer) & 65.93 & 29.51   &  18.88 \\
 & \kivifour{} & 66.38 & 29.49 & 23.65\\
 & \kivitwo{} & 66.23 & 29.84  & 20.77 \\
\midrule
\multirow{8}{*}{Falcon-7B}   & 16bit  &     59.83   &   23.20    &        4.55          \\
 & 4bit (K - $\mathbb{T}$, V - $\mathbb{T}$)  & 58.53 & 22.94 & 3.26\\
\cmidrule{2-5}
 & 2bit (K - $\mathbb{C}$, V - $\mathbb{T}$)   &  43.93 & 20.82 & 1.29 \\
 & 2bit (K - $\mathbb{T}$, V - $\mathbb{T}$)    & 25.72 & 0.91 & 0.53 \\
 & 2bit (K - $\mathbb{C}$, V - $\mathbb{C}$) & 41.95 & 17.11 & 1.52 \\
 & 2bit (K - $\mathbb{T}$, V - $\mathbb{C}$)   & 19.53 & 0.94 & 0.15 \\
 & \kivifour{} & 59.67 & 22.58 & 4.47 \\
 & \kivitwo{} & 57.48 & 24.98 & 3.41 \\
\midrule
\multirow{8}{*}{Mistral-7B}   & 16bit  &     67.40  &  30.45  &  38.36     \\
 & 4bit (K - $\mathbb{T}$, V - $\mathbb{T}$)  & 67.80 & 29.83 & 36.85 \\
\cmidrule{2-5}
 & 2bit (K - $\mathbb{C}$, V - $\mathbb{T}$)   & 61.65 & 29.64 & 26.46 \\
 & 2bit (K - $\mathbb{T}$, V - $\mathbb{T}$)     & 54.55 & 25.86 & 5.00 \\
 & 2bit (K - $\mathbb{C}$, V - $\mathbb{C}$) & 24.40 & 24.86 & 2.27  \\
 & 2bit (K - $\mathbb{T}$, V - $\mathbb{C}$)   & 10.73 & 19.12 & 0.99 \\
 % & kivi-2 (w/o buffer) & 66.45 & 29.76 & 28.73 \\
 & \kivifour{} & 66.95 & 30.49 & 37.30\\
 & \kivitwo{} & 66.35 & 32.17 & 36.01  \\
\bottomrule
\vspace{-1.5em}
\end{tabular}
}
\end{table}

\begin{table*}[t]
\centering
\caption{Performance evaluation of \kivi{} on various models across a range of benchmarks in LongBench. We highlight the average performance of our method.
\textbf{More similar results on Mistral-7B-v0.2 and LongChat-7b-v1.5 can be found in Table \ref{table:longchat} and Table \ref{table:mistral}}}
\label{tab:longbench}
\resizebox{\textwidth}{!}{
\begin{tabular}{llccccccccc} 
\toprule
\multicolumn{2}{l}{\textbf{Model}}    & \textbf{Qasper} & \textbf{QMSum} & \textbf{MultiNews} & \textbf{TREC} & \textbf{TriviaQA} & \textbf{SAMSum} & \textbf{LCC} & \textbf{RepoBench-P} & \textbf{Average}  \\ 
\midrule
\multirow{3}{*}{Llama2-7B}      & 16bit  & 9.52 & 21.28 & 3.51 & 66.00 & 87.72 & 41.69 & 66.66 & 59.82 & 44.52  \\
                                & \kivifour{} & 9.28 & 21.42 & 3.88 & 66.00 & 87.72 & 41.82 & 66.80 & 59.83 & 44.59\\
                                & \kivitwo{} & 9.31 & 20.50 & 1.14 & 66.00 & 87.42 & 42.71 & 66.88 & 60.23 & \textbf{44.27} \\
\midrule
\multirow{3}{*}{Llama2-13B}     & 16bit  & 9.32 & 21.38 & 3.71 & 70.00 & 87.87 & 43.55 & 66.61 &56.42 & 44.85 \\
                                & \kivifour{} & 9.16 & 20.86 & 3.21 & 69.00 & 86.97 & 44.26 & 65.30 & 57.08 & 44.48\\
                                & \kivitwo{} & 8.58 & 20.69 & 6.19 & 69.50 & 87.78 & 44.30 & 65.08 & 55.46 & \textbf{44.69} \\
\midrule
\multirow{3}{*}{Llama2-7B-Chat} & 16bit  & 19.65 & 20.54 & 26.36 & 63.00 & 84.28 & 41.12 & 59.75 & 52.93 & 45.95 \\
                                & \kivifour{} & 19.62 & 20.70 & 25.49 & 63.00 & 84.13 & 40.87 & 59.27 & 53.56 & 45.83\\
                                & \kivitwo{} & 19.32 & 20.46 & 25.48 & 63.00 & 84.84 & 40.60 & 58.71 & 52.97 & \textbf{45.67} \\
\midrule
\multirow{3}{*}{Llama2-13B-Chat} & 16bit  & 24.18 & 20.37 & 25.69 & 67.50 & 86.90 & 42.18 & 50.23 & 50.64 & 45.96 \\
                                & \kivifour{} & 23.00 & 20.36 & 26.06 & 67.50 & 87.20 & 42.04 & 52.55 & 52.77 & 46.44\\
                                 & \kivitwo{} & 23.59 & 20.76 & 25.25 & 67.5 & 87.17 & 41.56 & 49.93 & 48.45 & \textbf{45.52} \\
\midrule
\multirow{3}{*}{Falcon-7B}      & 16bit  & 1.48 & 2.35 & 11.09 & 13.00 & 5.84 & 2.44 & 23.86 & 9.69 & 8.71 \\
                                & \kivifour{} & 1.04 & 2.41 & 11.98 & 13.00 & 5.84 & 2.36 & 23.72 & 9.92 & \textbf{8.78} \\
                                & \kivitwo{} & 1.98 & 3.61 & 6.78 & 10.00 & 6.24 & 2.73 & 22.18 & 10.12 & \textbf{7.95} \\
\midrule
\multirow{3}{*}{Mistral-7B}     & 16bit  & 8.12 & 19.98 & 19.99 & 67.50 & 89.80 & 41.69 & 66.59 & 58.99 & 46.58 \\
                                & \kivifour{} & 7.89 & 20.06 & 20.58 & 67.50 & 89.80 & 41.56 & 66.45 & 58.62 & 46.56 \\
                                & \kivitwo{} & 6.92 & 19.71 & 17.92 & 66.50 & 89.63& 41.66 & 65.52 & 58.99 & \textbf{45.85} \\
\bottomrule
\end{tabular}
}
\end{table*}

\begin{figure*}[h]
% \vspace{-1em}
\setlength{\abovecaptionskip}{0mm}
\setlength{\belowcaptionskip}{0mm}
\centering
\subfigcapskip=-2mm

\begin{minipage}{0.95\linewidth}
    
\subfigure[\!Llama-3-8B-Instruct Baseline]{
\centering
    % \begin{minipage}[t]{0.3\linewidth}
        \includegraphics[width=0.3\linewidth]{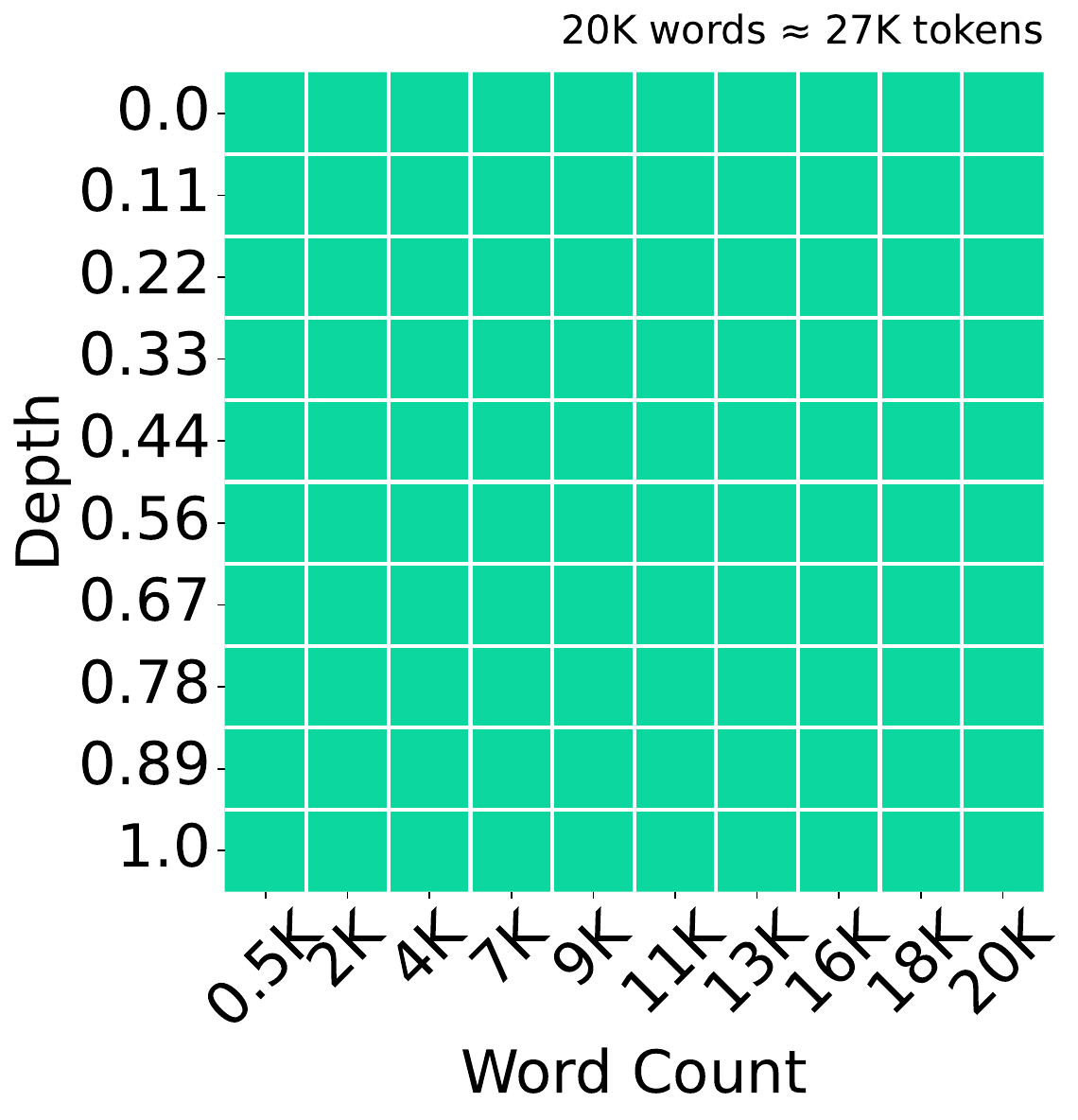}
    % \end{minipage}%
}
\subfigure[\!Llama-3-8B-Instruct + \kivitwo{}]{
\centering
    % \begin{minipage}[t]{0.3\linewidth}
        \includegraphics[width=0.3\linewidth]{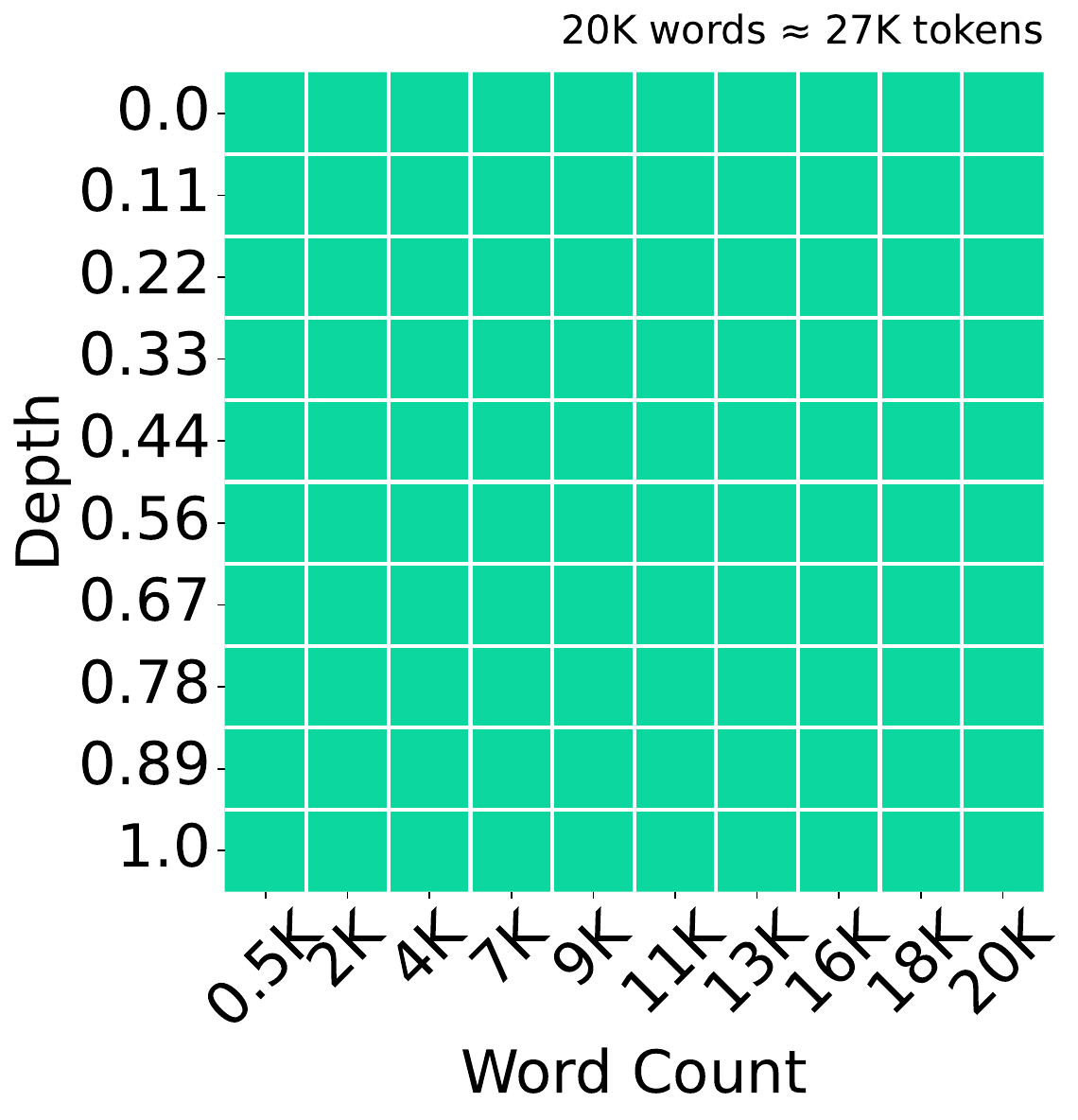}
    % \end{minipage}%
}
\subfigure[\!Llama-3-8B-Instruct + \kivifour{}]{
\centering
    % \begin{minipage}[t]{0.3\linewidth}
        \includegraphics[width=0.3\linewidth]{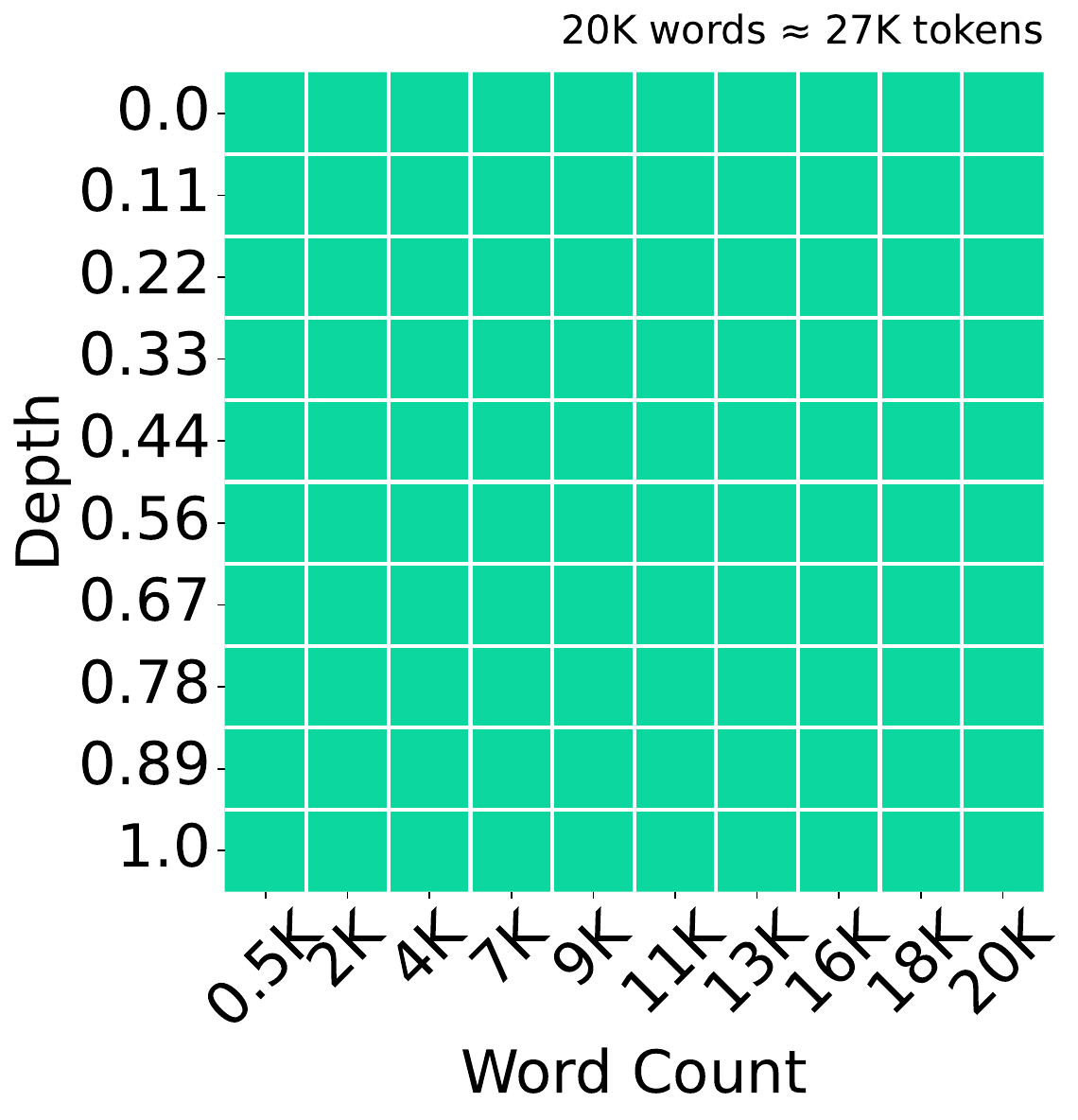}
    % \end{minipage}%
}

\subfigure[\!Mistral-7B-Instruct-v0.2 Baseline]{
\centering
    % \begin{minipage}[t]{0.3\linewidth}
        \includegraphics[width=0.3\linewidth]{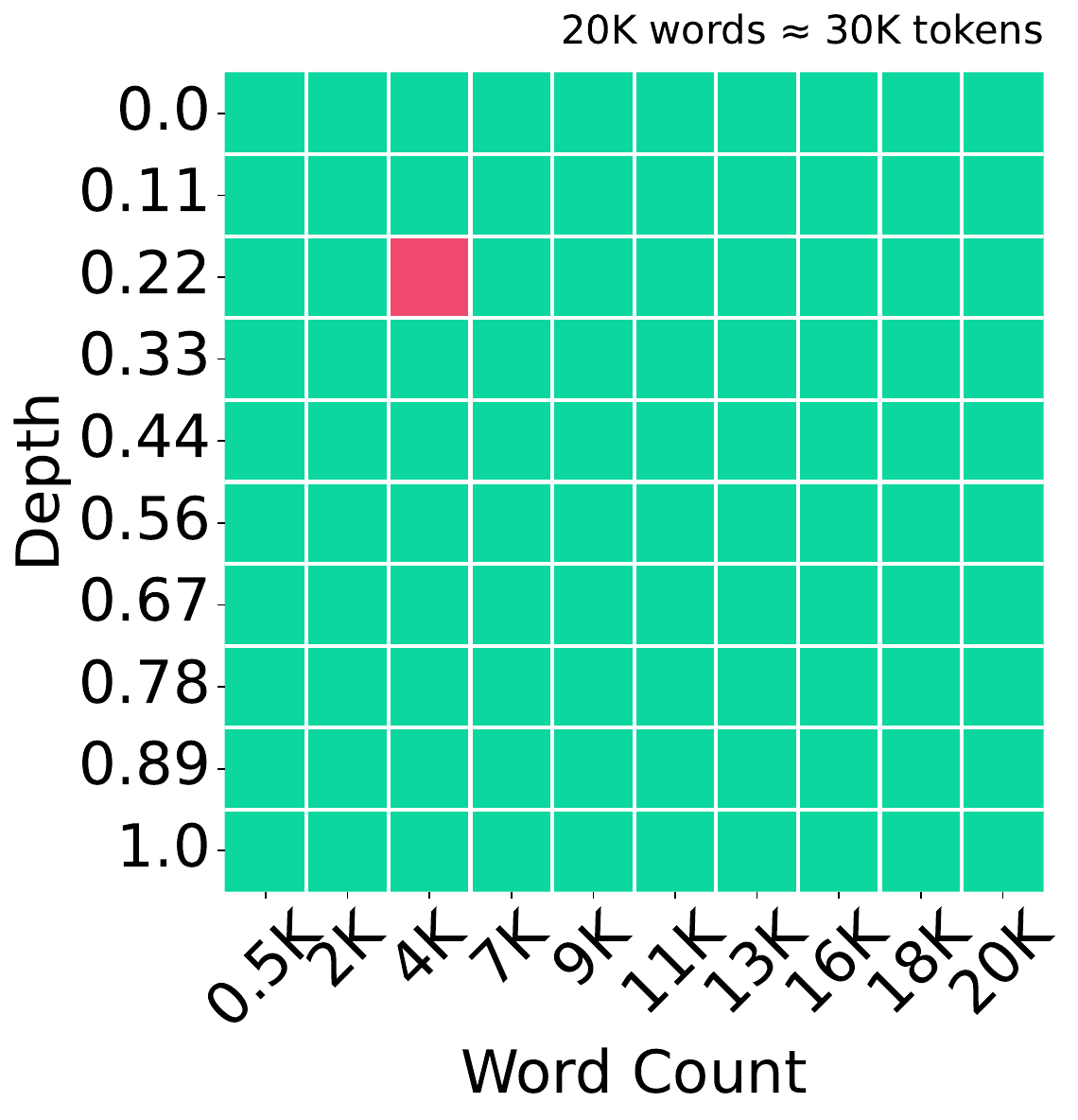}
    % \end{minipage}%
}
\subfigure[\!Mistral-7B-Instruct-v0.2 + \kivitwo{}]{
\centering
    % \begin{minipage}[t]{0.3\linewidth}
        \includegraphics[width=0.3\linewidth]{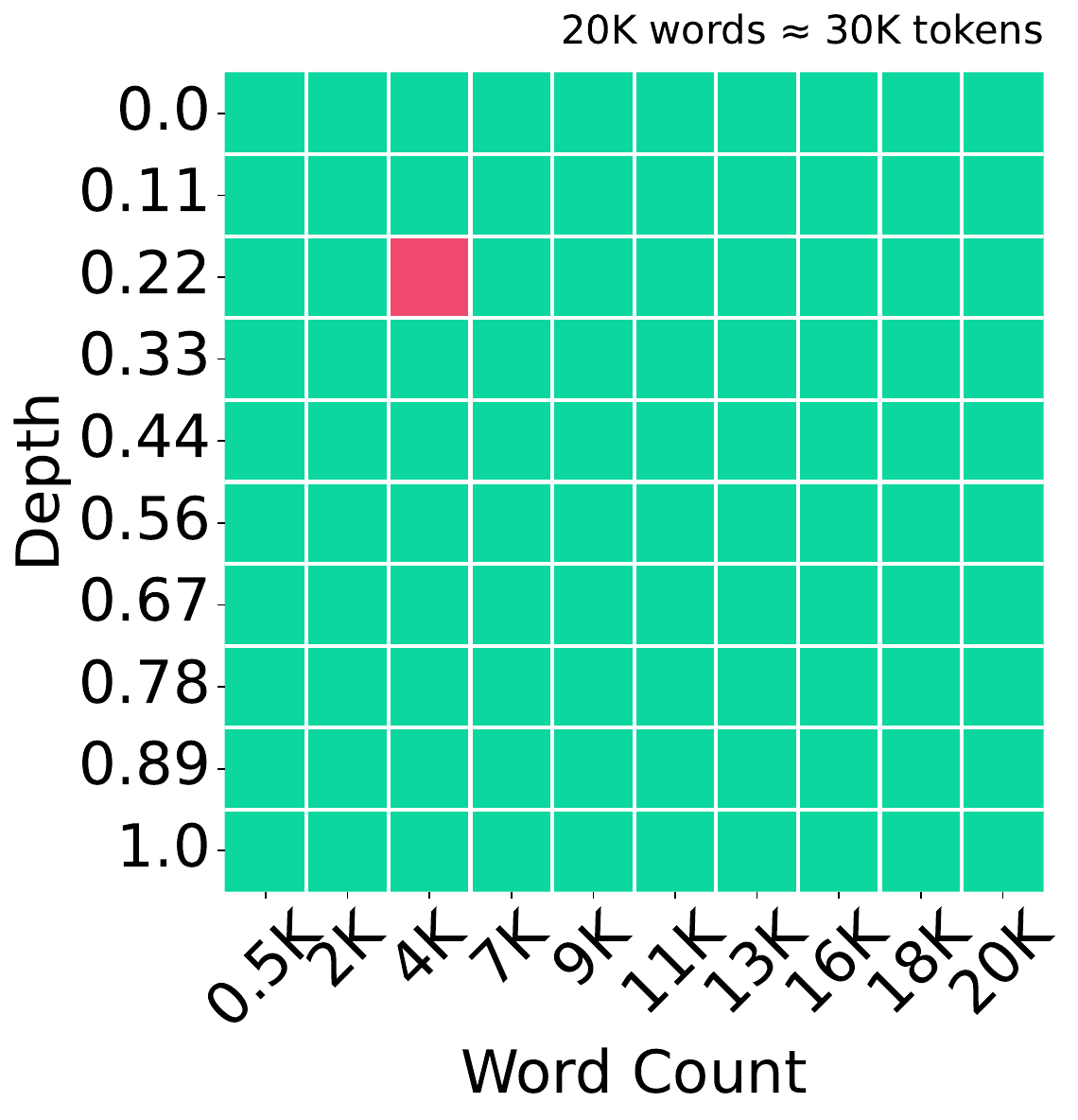}
    % \end{minipage}%
}
\subfigure[\!Mistral-7B-Instruct-v0.2 + \kivifour{}]{
\centering
    % \begin{minipage}[t]{0.3\linewidth}
        \includegraphics[width=0.3\linewidth]{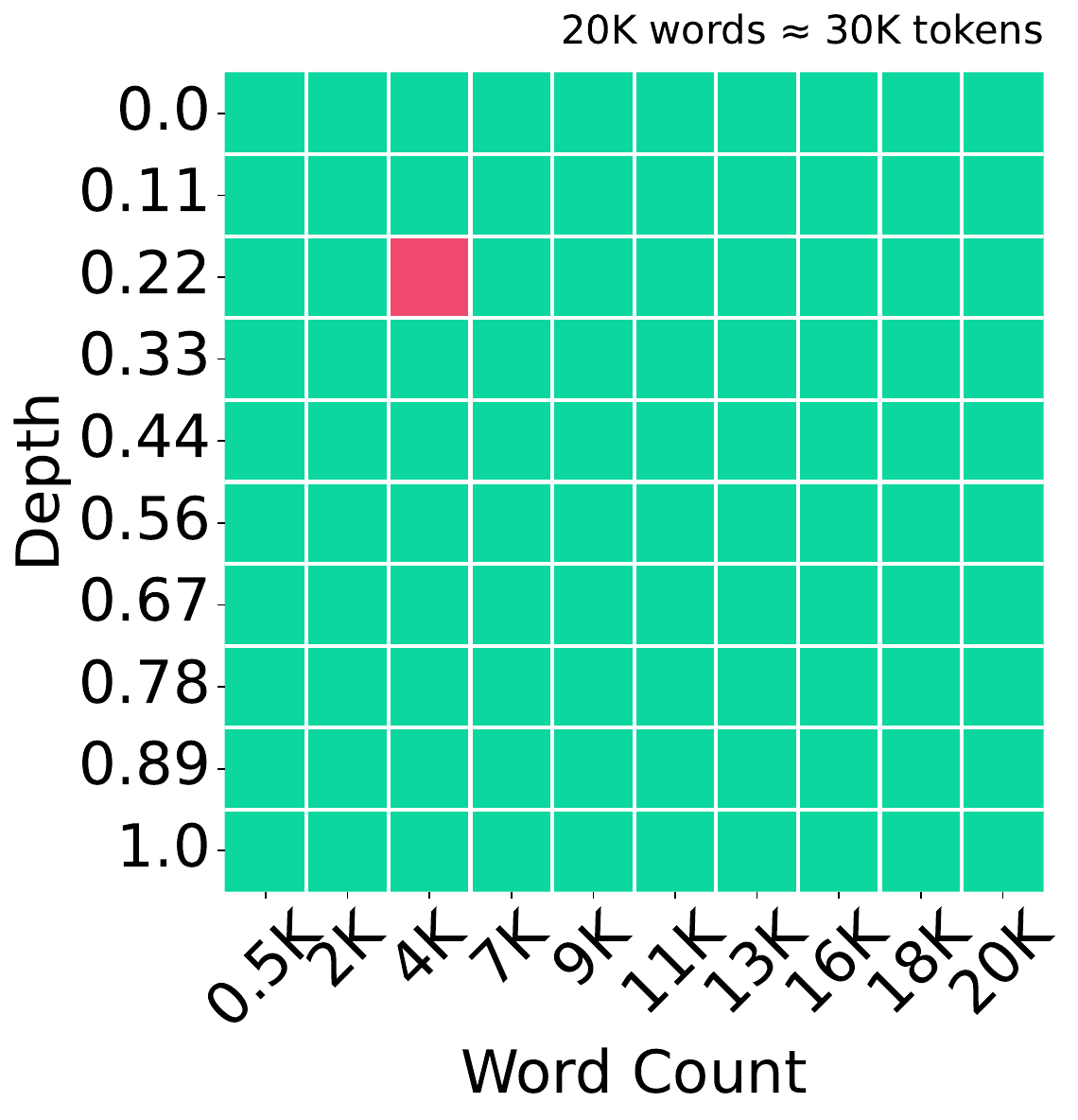}
    % \end{minipage}%
}

\end{minipage}
\hspace{-1em}
\begin{minipage}{0.04\linewidth}
    \centering
    \includegraphics[width=1.25\linewidth]{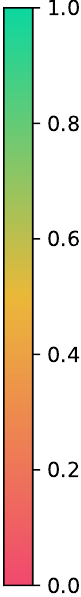}
\end{minipage}

\caption{Needle-in-a-Haystack results on Llama-3-8B-Instruct and Mistral-7B-Instruct-v0.2. Here we count the number of words instead of tokens to better account for the tokenizer difference. The final token length is noted in the upper right corners of each figure. Detailed setting can be found in Appendix \ref{app: niah}.}
\label{fig: needle}
\end{figure*}

\paragraph{LM-Eval Results.}
We benchmark \kivi{} in CoQA, TruthfulQA and GSM8K tasks using the LM-Eval framework. All dataset parameters were set to default. We compare the standard 16bit configuration with our \kivi{} compression techniques in Llama-2-7B, Llama-2-13B, Falcon-7B and Mistral-7B. As shown in Table ~\ref{tab:merged-lm-eval}, we observe that for the Llama and Mistral model, \kivi{} only has up to 2\% accuracy drop despite the KV cache being stored in 2bit.
For instance, in the Llama-2-7B model, the transition from 16bit to 2bit only slightly decreases accuracy. 
Similar trends are observed in other Llama-family models.
Since Falcon-7B adopts multi-query attention and only has one head for KV cache, it is already highly compressed compared to Llama-based models.
Thus, in Table \ref{tab:merged-lm-eval}, 4bit \kivi{} is needed to maintain the accuracy, while 2bit \kivi{} may have a large accuracy drop in this case.

\paragraph{LongBench Results.}
The performance of \kivi{} over various models in the LongBench dataset is summarised in Table ~\ref{tab:longbench}. We apply \kivi{} to Llama2-7B, Llama2-13B, Llama2-7B-Chat, Llama2-13B-Chat, Falcon-7B and Mistral-7B.
Table \ref{tab:longbench} suggests that \kivi{} is an effective method for KV cache compression with minimal impact on accuracy across various hard long context generation tasks.
\textbf{We present additional results using Llama3-8b, Mistral-7B-v0.2, and LongChat-7B-v1.5, which can be found in Appendix~\ref{app: exp}.}

\paragraph{NIAH Results.} From Figure \ref{fig: needle}, we observe that \kivi{} can still maintain the retrieval ability of LLMs even with 2bit KV Cache. 
Detailed NIAH setting can be found in Appendix \ref{app: niah}.

\begin{table}
\centering
\small
\caption{Ablation study of \kivi{} by changing group size $G$ and residual length $R$.}
% \vspace{-.5em}
\label{tab:abl}
\begin{tabular}{lcc} 
\toprule
\textbf{Model}                         & \textbf{Group Size} & \textbf{GSM8K}  \\ 
\midrule
\multirow{3}{*}{Llama2-13B}  & 32  & 20.77 \\
                            & 64  & 21.00 \\
                            & 128 & 17.29\\
\bottomrule
\toprule
\textbf{Model}                         & \textbf{Residual Length} & \textbf{GSM8K}  \\ 
\midrule
% \multirow{5}{*}{Llama2-13B}  & 0  & 18.88 \\ 
 \multirow{4}{*}{Llama2-13B}                            & 32  & 20.62 \\
                             & 64  & 19.86 \\
                             & 96  & 20.55\\
                             & 128 & 20.77\\
\bottomrule
\end{tabular}
\vspace{-1.5em}
\end{table} 

\subsubsection{Ablation}

In this section, we benchmark \kivi{} on GSM8K, one of the hardest generation tasks, to show the effect of hyperparameters group size $G$ and residual length $R$ on the model performance. For full results of \kivi{} with a residual length of 32, please refer to Appendix~\ref{app: abl}.

\textbf{The effect of group size.} We fix the residual length at 128 and vary the group sizes to 32, 64, and 128. From Table~\ref{tab:abl}, we observe that group sizes 32 and 64 yield similar results, whereas the performance significantly decreases when the group size reaches 128. Note the zero-point and the scaling factor mentioned in Section~\ref{sec: quant prelim} are calculated according to this group size; where the choice of group size will greatly impact the KV cache compression effect under a long input.

\textbf{The effect of residual length.} We fix the group size at 32 and vary the residual length across 32, 64, 96, and 128. As shown in Table~\ref{tab:abl}, 
there is no consistent pattern between residual lengths and model accuracy.
Namely, while a residual length of 128 achieves good results, 32 and 96 yield similar outcomes, but a residual length of 64 results in the worst performance. We emphasize that while we observe no significance among residual lengths of $\{32, 96, 128\}$, \textbf{having a reasonably large residual length is important; as it brings much performance boosts on hard tasks like GSM8K, again as shown in Table~\ref{tab:abl}.}

% the performance associated with different residual lengths can be arbitrary; while a residual length of 128 achieves good results, 32 and 96 yield similar outcomes, but a residual length of 64 results in the worst performance.

% As Section~\ref{sec: efficiency} showed, the residual length is important to the efficiency metric. 

\begin{figure}
 % \vspace{-4em}
  \begin{center}
    \includegraphics[width=0.39\textwidth]{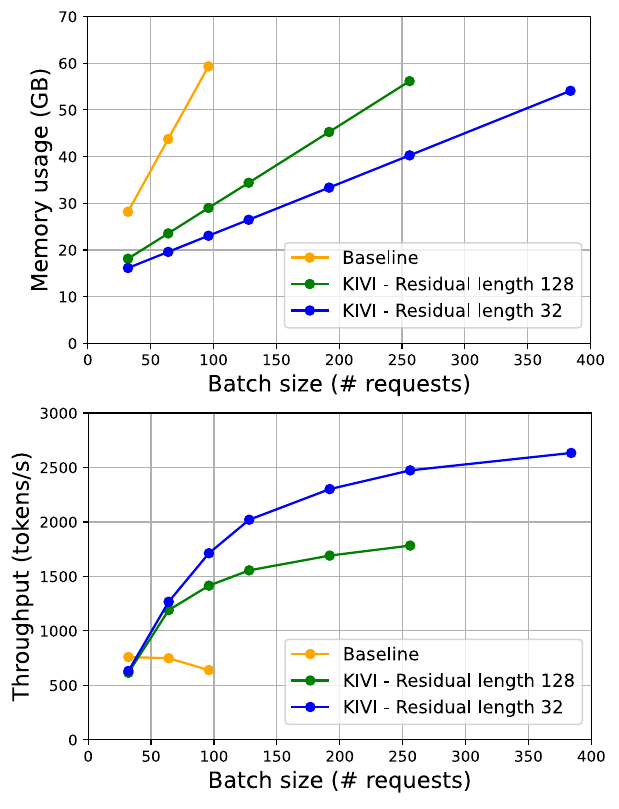}
  \end{center}
    \vspace{-1em}
  \caption{Memory usage and throughput comparison between 2bit \kivi{} and 16bit baseline. \kivi{} can achieve higher throughput by enabling a larger batch size.}
  \label{fig: throughput}
\end{figure}

\subsubsection{Efficiency Comparison}
\label{sec: efficiency}

To evaluate the wall-clock time efficiency of \kivi{},
following vLLM \citep{vllm}, we synthesize workloads based on ShareGPT \citep{shargpt}, which contain input and output
texts of real LLM services.
On average, the data set has an input prompt length $l_{\text{prompt}}$ of 161 and an output length $l_{\text{gen}}$ of 338 \citep{vllm}.
We increase the batch size until out of memory and report the peak memory usage and throughput between \kivi{} (with residual length 32 and 128) and FP16 baseline for the Llama-2-7B model.
The hardware here is a single NVIDIA A100 GPU (80GB).

As shown in Figure \ref{fig: throughput}, with similar maximum memory usage, \kivi{} enables up to $\mathbf{4\times}$ larger batch size and gives $\mathbf{2.35\times \sim 3.47\times}$ larger throughput.
This throughput number can grow larger with longer context length and output length.
We also note that \textbf{ this speed-up can be greatly increased if we further fuse the KV cache quantization process with previous operations.}
We leave it as one of future work.
% Besides the ShareGPT workload, we also report the memory and speed of \kivi with longer context and output length on synthesized workload.

\section{Related Work}

Many machine learning systems and benchmark works consider scaling up LLM inference process \citep{pope2023efficiently, yuan2024kv}.
Among them, quantization techniques have been widely applied \citep{gptq, lin2023awq, kim2023squeezellm, xu2023compress}. A main branch of LLM quantization is weight-only quantization, which involves the quantization of model weights to lower precision. For instance, AWQ~\citep{lin2023awq} cleverly quantizes model weights to INT4 and INT3 using an activation-aware manner. GPTQ~\citep{gptq} utilizes approximate second-order information to quantize model weights both accurately and efficiently. SqueezeLLM~\citep{kim2023squeezellm} adopts the concept of non-uniform quantization based on sensitivity along with dense-and-sparse decomposition. 
This line of work is orthogonal to ours, as they can be combined.

SmoothQuant~\citep{xiao2023smoothquant} is a post-training quantization method that is more closely related to our work. This method uses equivalent transformations to balance the quantization complexity for both activation and weight, making the activation easier to quantize. SmoothQuant can compress KV cache to 8bit with minor performance loss. However, it faces a significant accuracy drop when scaled down to 4bit or less~\citep{zhao2024atom}.
FlexGen~\citep{flexgen} adopts 4-bit group-wise quantization for both key and value cache.

One essential recipe of \kivi{} is the per-channel quantization scheme designed based on the observation made in Section~\ref{sec: analysis} and Figure~\ref{fig: vis}. 
ATOM \citep{zhao2024atom} also indicates that key cache exhibits more outliers compared to the value cache. \kivi{} provides further extensive analysis and leverages this observation to implement per-channel quantization. A similar observation and approach has been independently discovered and developed in the concurrent work KVQuant \citep{hooper2024kvquant}.

vLLM \citep{vllm} and S3 \citep{jin2023s} are system-level works, which include memory management through the use of PagedAttention or memory usage prediction. They can lower the memory requirements of KV cache and simultaneously increase model throughput. This research direction is orthogonal to our work, since system-level optimizations can also be applied upon our algorithm.

Several other works also consider compressing KV cache by evicting tokens. 
H2O~\citep{h2o} retains only a small portion of tokens that contribute significantly to the attention scores. Similarly, Scissorhands~\citep{liu2024scissorhands} exploit the persistence of the importance hypothesis in KV cache sparsification.  
StreamingLLM~\citep{xiao2023efficient} is based on the observation of ``attention sink'' and maintains only a few initial tokens to preserve performance. Unlike these works, our \kivi{} retains all input tokens and compresses them into lower precision. This line of work is orthogonal to ours, as they can also be combined together.

\section{Conclusion and Future Work}

In this paper, we systematically analyze KV cache element distribution in popular LLMs.
We conclude that key cache should be quantized per-channel and value cache should be quantized per token.
Based on these observations, we propose \kivi{}, a plug-and-play 2bit KV cache quantization algorithm without the need for any tuning. In real LLM workload, \kivi{} allows up to $\mathbf{4\times}$ larger batch sizes and $\mathbf{3.47\times}$ throughput. 
In the future, we will further optimize the implementation to reduce the overhead of quantization process during the prefill and decoding phase.

\section*{Acknowledgments}

The authors thank the anonymous reviewers for their helpful comments. 
Dr. Beidi Chen is supported by a research gift from Moffett.AI.
Dr. Vladimir Braverman is partially supported by the Ministry of Trade, Industry and Energy (MOTIE) and Korea Institute for Advancement of Technology (KIAT) through the International Cooperative R\&D program, the Naval Research (ONR) grant N00014-23-1-2737, and NSF CNS 2333887 award.
Dr. Xia Hu is supported by NSF grants NSF IIS-2224843. 
The views and conclusions contained in this paper are those of the authors and should not be interpreted as representing any funding agencies.

\section*{Impact Statement}

This paper presents work whose goal is to advance the field of Machine Learning. There are many potential societal consequences of our work, none which we feel must be specifically highlighted here.

\bibliographystyle{plainnat}
\bibliography{ref}

\clearpage
\appendix
\onecolumn
\section{Detailed Implementations}
\label{app: implement}

In this section, we present the algorithm for \kivi{} as discussed in Section~\ref{sec: algo}. Specifically, we provide the pseudocode for \kivi{} when calculating the attention output in the prefill and decoding phases.

\begin{algorithm}[h!]  
 \SetKwInput{KwParam}{parameter}
 \KwParam{group size $G$, residual length $R$}
  \SetKwProg{myProcedure}{procedure}{\string:}{end procedure}
  \SetKwProg{myFunc}{function}{\string:}{end function}
  \myProcedure{\FuncSty{Prefill}}{
\KwIn{$\mX\in{\mathbb{R}^{l_{\text{prompt}}\times d}}$}
  $\mX_K = \mX \mW_K, \mX_V = \mX \mW_V$ \\
    $\mX_{V_g}=\mX_V[:l_{\text{prompt}}-R], \mX_{V_r}=\mX_V[l_{\text{prompt}}-R:]$ \\
    $Q(\mX_{V_g}) \leftarrow$ GroupQuant$(\mX_{V_g}, \text{dim=token}, \text{numGroup=$d//G$})$ \\
    $Q(\mX_{K_g}), \mX_{K_r} \leftarrow \FuncSty{KeyQuant}(\mX_K)$ \\
    \texttt{KV cache} $\leftarrow Q(\mX_{K_g}), \mX_{K_r}, Q(\mX_{V_g}), \mX_{V_r}$ \\
    \Return{$\mX_K, \mX_V$}
  }
    \myProcedure{\FuncSty{Decoding}}{
      \KwIn{\texttt{KV cache}, $\vt\in{\mathbb{R}^{1\times d}}$}
$\vt_Q = \vt \mW_Q, \vt_K = \vt \mW_K, \vt_V = \vt \mW_V$ \\
$Q(\mX_{K_g}), \mX_{K_r}, Q(\mX_{V_g}), \mX_{V_r} \leftarrow\texttt{KV cache}$ \\ 
$\mX_{K_r}\leftarrow \text{Concat}([\mX_{K_r}, \vt_K], \text{dim=token})$\\
$\mX_{V_r}\leftarrow \text{Concat}([\mX_{V_r}, \vt_V], \text{dim=token})$\\
\If{$\text{len}(\mX_{K_r})=R$}{
    $Q(\mX_{K_r}), \_ \leftarrow$\FuncSty{KeyQuant}($\mX_{K_r}$)\\
    $Q(\mX_{K_g})\leftarrow \text{Concat}([Q(\mX_{K_g}), Q(\mX_{K_r})], \text{dim=token})$ \\
    $\mX_{K_r}\leftarrow$ empty tensor.
}
\If{$\text{len}(\mX_{V_r})>R$}{
    $Q(\mX_{V_r'}) \leftarrow$ GroupQuant$(\mX_{V_r}[:-R], \text{dim=token}, \text{numGroup}=d//G)$ \\
    $Q(\mX_{V_g})\leftarrow \text{Concat}([Q(\mX_{V_g}), Q(\mX_{V_r'})], \text{dim=token})$ \\
    $\mX_{V_r} \leftarrow \mX_{V_r}[-R:]$
}
$\mA \leftarrow \text{Concat}([\vt_Q Q(\mX_{K_g})^\top, \vt_Q \mX_{K_r}^\top], \text{dim=token})$\\
$\mA_g = \text{Softmax}(\mA)[:-R], \mA_r = \text{Softmax}(\mA)[-R:]$ \\
$\vt_O \leftarrow \mA_g Q(\mX_{V_g}) + \mA_r \mX_{V_r}$  \\
\texttt{KV cache} $\leftarrow Q(\mX_{K_g}), \mX_{K_r}, Q(\mX_{V_g}), \mX_{V_r}$ \\
\Return{$\vt_O$}
  }
  
    \myFunc{\FuncSty{KeyQuant}($\mX_K\in\mathbb{R}^{l\times d}$)}{
  $ r = l \% R$, \\
  $\mX_{K_g}=\mX_K[:l-r], \mX_{K_r}=\mX_K[l-r:]$ \\
  $Q(\mX_{K_g}) \leftarrow$ GroupQuant$(\mX_{K_g}, \text{dim=channel}, \text{numGroup=$l // G$})$ \\
  \Return{$Q(\mX_{K_g}), \mX_{K_r}$}
  }
  \caption{The \kivi{} Prefill \& Decoding Algorithm}
  \label{algo: kivi}
\end{algorithm}

\newpage
\section{NIAH Setting}
\label{app: niah}

We largely follows the passkey retrieval prompt template of \citet{mohtashami2023passkey} but using 7-digit passkey and Paul Graham Essays\footnote{\url{https://paulgraham.com/articles.html}} as the background filler, as set forth in \href{https://github.com/gkamradt/LLMTest_NeedleInAHaystack}{Arize-ai} and \citet{reid2024gemini}:

\noindent {\small\texttt{There is an important info hidden inside a lot of irrelevant text.  Find it and memorize them.  I will quiz you about the important information there.}}

\noindent {\small\texttt{<prefix filled by Paul Graham Essays>}}

\noindent {\small\texttt{The pass key is <7-DIGIT PASS KEY>.  Remember it. <7-DIGIT PASS KEY> is the pass key.}}

\noindent {\small\texttt{<suffix filler>}}

\noindent {\small\texttt{What is the pass key?  The pass key is}}

\section{More Ablation Results}
\label{app: abl}

In our efficiency evaluation, we observe that with a residual length of 32, \kivi{} achieves a significantly higher memory compression rate, which in turn leads to increased throughput. Additionally, our ablation study reveals that changing the residual length from 128 to 32 does not result in a substantial performance gap. We demonstrate \kivi{} with a residual length of 32 across all benchmark datasets. As shown in Tables~\ref{tab: abl lmeval} and~\ref{tab: abl longbench}, \kivi{} with a residual length of 32 also delivers performance comparable to that of the 16-bit full model.

\begin{table}[h!]
\centering
\caption{Performance comparison between 16bit, \kivitwo{} (2bit) / \kivifour{} (4bit) with residual length 128 and 32 across various models. $R32$ stands for residual length 32.}
\label{tab: abl lmeval}
\resizebox{0.6\linewidth}{!}{
\begin{tabular}{llccc} 
\toprule
\multicolumn{2}{l}{\textbf{Model}}        & \textbf{CoQA} & \textbf{TruthfulQA} & \textbf{GSM8K}  \\
\midrule
\multirow{3}{*}{Llama-2-7B}  & 16bit  &     63.88      &         30.76        &   13.50     \\
& \kivitwo{} $R128$ &  63.05  & 33.95  &   12.74 \\
& \kivitwo{} $R32$ &   62.85 & 33.01 & 13.57 \\
\midrule
\multirow{3}{*}{Llama-2-13B} & 16bit  &      66.37     &        29.53        &   22.67      \\
 & \kivitwo{} $R128$ & 66.23 & 29.84  & 20.77 \\
 & \kivitwo{} $R32$ & 66.57 & 29.35 & 20.62\\
\midrule
\multirow{5}{*}{Falcon-7B}   & 16bit  &     59.83   &   23.20    &        4.55          \\
 & \kivifour{} $R128$ &                         59.67 & 22.58 & 4.47 \\
 & \kivifour{} $R32$ &                         59.73 & 22.96 & 3.94\\
 & \kivitwo{} $R128$ &                              57.48 & 24.98 & 3.41 \\
 & \kivitwo{} $R32$ &                          57.50 & 25.70 & 2.20 \\
\midrule
\multirow{3}{*}{Mistral-7B}   & 16bit  &     67.40  &  30.45  &  38.36     \\
 & \kivitwo{} $R128$ & 66.35 & 32.17 & 36.01  \\
 & \kivitwo{} $R32$ &  65.90 & 31.21 & 34.34\\
\bottomrule

\end{tabular}
}
\end{table}

\begin{table*}[h!]
\centering
\caption{Performance evaluation of \kivi{} with residual length 128 and 32 on various models across a range of benchmarks in LongBench. $R32$ stands for residual length 32.}
\label{tab: abl longbench}
\resizebox{\textwidth}{!}{
\begin{tabular}{llccccccccc}
\toprule
\multicolumn{2}{l}{\textbf{Model}}    & \textbf{Qasper} & \textbf{QMSum} & \textbf{MultiNews} & \textbf{TREC} & \textbf{TriviaQA} & \textbf{SAMSum} & \textbf{LCC} & \textbf{RepoBench-P} & \textbf{Average}  \\ 
\midrule
\multirow{3}{*}{Llama2-7B}      & 16bit  & 9.52 & 21.28 & 3.51 & 66.00 & 87.72 & 41.69 & 66.66 & 59.82 & 44.52  \\
                                & \kivitwo{} $R128$ & 9.31 & 20.50 & 1.14 & 66.00 & 87.42 & 42.71 & 66.88 & 60.23 & 44.27 \\
                                & \kivitwo{} $R32$ & 9.26 & 20.53 & 0.97 & 66.00 & 87.42 & 42.61 & 66.22 & 59.67 & 44.08\\
\midrule
\multirow{3}{*}{Llama2-13B}     & 16bit  & 9.32 & 21.38 & 3.71 & 70.00 & 87.87 & 43.55 & 66.61 &56.42 & 44.85 \\
                                & \kivitwo{} $R128$ & 8.58 & 20.69 & 6.19 & 69.50 & 87.78 & 44.30 & 65.08 & 55.46 & 44.69 \\
                                & \kivitwo{} $R32$ & 8.38 & 20.74 & 7.01 & 69.50 & 87.78 & 44.43 & 64.89 & 55.31 & 44.75 \\
\midrule
\multirow{3}{*}{Llama2-7B-Chat} & 16bit  & 19.65 & 20.54 & 26.36 & 63.00 & 84.28 & 41.12 & 59.75 & 52.93 & 45.95 \\
                                & \kivitwo{} $R128$ & 19.32 & 20.46 & 25.48 & 63.00 & 84.84 & 40.60 & 58.71 & 52.97 & 45.67 \\
                                & \kivitwo{} $R32$ & 19.10 & 20.08 & 25.33 & 63.00 & 85.04 & 39.80 & 57.91 & 52.38 & 45.33 \\
\midrule
\multirow{3}{*}{Llama2-13B-Chat} & 16bit  & 24.18 & 20.37 & 25.69 & 67.50 & 86.90 & 42.18 & 50.23 & 50.64 & 45.96 \\
                                 & \kivitwo{} $R128$ & 23.59 & 20.76 & 25.25 & 67.50 & 87.17 & 41.56 & 49.93 & 48.45 & 45.52 \\
                                 & \kivitwo{} $R32$ & 23.56 & 20.90 & 25.45 & 67.50 & 87.42 & 41.40 & 48.93 & 48.81 & 45.49 \\
\midrule
\multirow{5}{*}{Falcon-7B}      & 16bit  & 1.48 & 2.35 & 11.09 & 13.00 & 5.84 & 2.44 & 23.86 & 9.69 & 8.71 \\
                                & \kivifour{} $R128$ & 1.04 & 2.41 & 11.98 & 13.00 & 5.84 & 2.36 & 23.72 & 9.92 & 8.78 \\
                                & \kivifour{} $R32$ & 1.03 & 2.45 & 11.99 & 13.50 & 5.84 & 2.46 & 23.88 & 9.95 & 8.88 \\
                                & \kivitwo{} $R128$ & 1.98 & 3.61 & 6.78 & 10.00 & 6.24 & 2.73 & 22.18 & 10.12 & 7.95 \\
                                & \kivitwo{} $R32$ & 2.28 & 3.23 & 6.73 & 10.00 & 6.31 & 2.88 & 22.71 & 10.45 & 8.07 \\
\midrule
\multirow{3}{*}{Mistral-7B}     & 16bit  & 8.12 & 19.98 & 19.99 & 67.50 & 89.80 & 41.69 & 66.59 & 58.99 & 46.58 \\
                                & \kivitwo{} $R128$ & 6.92 & 19.71 & 17.92 & 66.50 & 89.63 & 41.66 & 65.52 & 58.99 & 45.85 \\
                                & \kivitwo{} $R32$ & 6.84 & 19.81 & 17.20 & 66.50 & 89.63 & 42.82 & 65.13 & 58.06 & 45.74 \\
\bottomrule
\end{tabular}
}
\end{table*}

\section{More Experimental Results}
\label{app: exp}

We present additional results using Llama3-8B, Mistral-7B-v0.2, and LongChat-7B-v1.5 in LongBench, which can be found in Table~\ref{table:llama3}, Table~\ref{table:mistral} and Table~\ref{table:longchat}, respectively.

We also show result of Needle-in-a-Haystack Test in Figure~\ref{fig: needle}. The settings largely follow the format of the original passkey retrieval task \citep{mohtashami2023passkey} while including some modern modifications set forward by \href{https://github.com/gkamradt/LLMTest_NeedleInAHaystack}{Arize-ai} and the technical report of Gemini 1.5 \citep{reid2024gemini}.

\begin{table*}[h]
\centering
\caption{The results of Llama-3-8B-Instruct with \kivi{} on LongBench.
The model has 8K context length and applies group query attention, which uses 8 heads for KV cache instead of the full 32 heads. We use a 32 group size and 128 residual length for both \kivitwo{} and \kivifour{}. The baseline is of full precision.}
\resizebox{\textwidth}{!}{
\begin{tabular}{lcccccccc}
\toprule
 &  \textbf{NarrativeQA} & \textbf{Qasper} & \textbf{MultiFieldQA} & \textbf{HotpotQA} & \textbf{MuSiQue} & \textbf{2WikiMQA} & \textbf{GovReport} & \textbf{QMSum} \\
\midrule
Baseline & 21.71 & 44.24 & 44.54 & 46.82 & 21.49 & 36.42 & 30.03 & 22.67 \\
w./ \kivitwo{} & 21.35 & 43.17 & 44.49 & 46.79 & 20.56 & 37.05 & 29.98 & 22.07  \\
w./ \kivifour{} & 21.01 & 44.83 & 44.60 & 46.96 & 21.43 & 36.48 & 30.22 & 22.44  \\ 
\midrule
 & \textbf{MultiNews} & \textbf{LCC} & \textbf{RepoBench-P} & \textbf{TriviaQA} & \textbf{SAMSum} & \textbf{TRec} & \textbf{PR} & \textbf{Avg} \\
\midrule
Baseline & 27.79 & 57.00 & 51.22 & 90.23 & 42.53 & 74.50 & 67.00 & 45.21 \\
w./ \kivitwo{} & 27.77 & 50.84 & 46.65 & 90.54 & 42.26 & 74.50 & 67.50 & 44.37 \\
w./ \kivifour{} & 27.97 & 57.36 & 52.03 & 90.33 & 42.97 & 74.50 & 66.50 & 45.31 \\ 
\bottomrule
\end{tabular}
}
\label{table:llama3}
\end{table*}

\begin{table*}[h]
\centering
\caption{The results of Mistral-7B-Instruct-v0.2 with \kivi{} on LongBench.
The model has 32K context length and applies group query attention, which uses 8 heads for KV cache instead of the full 32 heads. We use a 32 group size and 128 residual length for both \kivitwo{} and \kivifour{}. The baseline is of full precision.}
\resizebox{\textwidth}{!}{
\begin{tabular}{lcccccccc}
\toprule
 &  \textbf{NarrativeQA} & \textbf{Qasper} & \textbf{MultiFieldQA} & \textbf{HotpotQA} & \textbf{MuSiQue} & \textbf{2WikiMQA} & \textbf{GovReport} & \textbf{QMSum} \\
\midrule
Baseline & 21.02 & 29.41 & 47.13 & 36.53 & 19.13 & 21.76 & 32.59 & 23.99 \\
w./ \kivitwo{} & 20.61 & 28.73 & 44.88 & 35.47 & 17.95 & 20.68 & 32.55 & 23.65 \\
w./ \kivifour{} & 20.97 & 29.41 & 46.52 & 36.25 & 19.53 & 21.66 & 32.97 & 24.06 \\ 
\midrule
 & \textbf{MultiNews} & \textbf{LCC} & \textbf{RepoBench-P} & \textbf{TriviaQA} & \textbf{SAMSum} & \textbf{TRec} & \textbf{PR} & \textbf{Avg} \\
\midrule
Baseline & 27.09 & 53.49 & 51.40 & 86.23 & 43.04 & 71.00 & 89.33 & 43.54 \\
w./ \kivitwo{} & 26.54 & 53.03 & 51.16 & 86.00 & 43.34 & 71.00 & 80.83 & 42.43 \\
w./ \kivifour{} & 26.89 & 53.33 & 51.41 & 86.23 & 43.34 & 71.00 & 89.42 & 43.53 \\ 
\bottomrule
\end{tabular}
}
\label{table:mistral}
\end{table*}

\begin{table*}[h]
\centering
\caption{The results of LongChat-7B-v1.5-32K with \kivi{} on LongBench.
The model has 32K context length. We use a 32 group size and 128 residual length for both \kivitwo{} and \kivifour{}. The baseline is of full precision.}
\resizebox{\textwidth}{!}{
\begin{tabular}{lcccccccc}
\toprule
  &  \textbf{NarrativeQA} & \textbf{Qasper} & \textbf{MultiFieldQA} & \textbf{HotpotQA} & \textbf{MuSiQue} & \textbf{2WikiMQA} & \textbf{GovReport} & \textbf{QMSum} \\ 
\midrule
Baseline & 20.65 & 29.42 & 43.15 & 33.05 & 14.66 & 24.14 & 30.85 & 22.84 \\
w./ \kivitwo{} & 20.79 & 28.69 & 41.02 & 32.91 & 13.82 & 23.00 & 30.47 & 22.59 \\
w./ \kivifour{} & 20.49 & 28.90 & 43.24 & 33.07 & 14.66 & 24.86 & 31.40 & 22.84 \\ 
\midrule
 & \textbf{MultiNews} & \textbf{LCC} & \textbf{RepoBench-P} & \textbf{TriviaQA} & \textbf{SAMSum} & \textbf{TRec} & \textbf{PR} & \textbf{Avg} \\ 
\midrule
Baseline & 26.55 & 54.83 & 58.94 & 83.99 & 40.75 & 66.50 & 30.50 & 38.72 \\
w./ \kivitwo{} & 26.28 & 54.11 & 57.62 & 83.19 & 41.28 & 66.50 & 32.25 & 38.30 \\
w./ \kivifour{} & 26.52 & 54.06 & 58.77 & 83.88 & 40.62 & 67.00 & 31.50 & 38.79 \\ 
\bottomrule
\end{tabular}
}
\label{table:longchat}
\end{table*}

\end{document}